\documentclass{article}

\usepackage{PRIMEarxiv}

\usepackage[utf8]{inputenc} %
\usepackage[T1]{fontenc}    %
\usepackage{hyperref}       %
\usepackage{url}            %
\usepackage{booktabs}       %
\usepackage{amsfonts}       %
\usepackage{nicefrac}       %
\usepackage{microtype}      %
\usepackage{lipsum}
\usepackage{fancyhdr}       %
\usepackage{graphicx}       %
\graphicspath{{media/}}     %

\usepackage{latexsym}
\usepackage{multicol,multirow}
\usepackage{amsmath,amssymb,amsfonts}
\usepackage{mathrsfs}
\usepackage{amsthm}
\usepackage{rotating}
\usepackage{appendix}
\usepackage[authoryear]{natbib}
\usepackage{ifpdf}
\usepackage[T1]{fontenc}
\usepackage{times}
\usepackage{newtxmath}
\usepackage{textcomp}%
\usepackage{xcolor}%
\usepackage{hyperref}
\usepackage{lipsum}
\usepackage{cleveref}
\usepackage{float}
\usepackage{subcaption}
\newcommand{\R}{{\mathbb{R}}}
\title{Precipitation nowcasting of satellite data using physically-aligned neural networks
}
\author{ \href{https://orcid.org/0009-0004-7132-8481}{\includegraphics[scale=1]{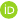}}\hspace{1mm}Antônio Catão \thanks{Instituto Nacional de Matemática Pura e Aplicada, Rio de Janeiro, RJ, Brazil}\\
  antonio.catao@impa.br\\
  \And
  \href{https://orcid.org/0009-0006-8144-0666}{\includegraphics[scale=1]{orcid_logo-eps-converted-to.pdf}}\hspace{1mm} Melvin Poveda\footnotemark[\value{footnote}]\\
  melvin.poveda@impa.br
  \\
  \And
  \href{https://orcid.org/0009-0008-2990-5969}{\includegraphics[scale=1]{orcid_logo-eps-converted-to.pdf}}\hspace{1mm}Leonardo Voltarelli\footnotemark[\value{footnote}]\\
  leonardo.voltarelli@impa.br\\
  \And
  \href{https://orcid.org/0000-0003-0907-704X}{\includegraphics[scale=1]{orcid_logo-eps-converted-to.pdf}}\hspace{1mm}Paulo Orenstein\footnotemark[\value{footnote}]\\
  pauloo@impa.br\\
}

\begin{document}
\maketitle

\begin{abstract}
  Accurate short-term precipitation forecasts predominantly rely on dense weather-radar networks, limiting operational value in places most exposed to climate extremes. We present TUPANN (Transferable and Universal Physics-Aligned Nowcasting Network), a satellite-only model trained on GOES-16 RRQPE.
  Unlike most deep learning models for nowcasting, TUPANN decomposes the forecast into physically meaningful components: a variational encoder–decoder infers motion and intensity fields from recent imagery under optical-flow supervision, a lead-time-conditioned MaxViT evolves the latent state, and a differentiable advection operator reconstructs future frames. We evaluate TUPANN on both GOES-16 and IMERG data, in up to four distinct climates (Rio de Janeiro, Manaus, Miami, La Paz) at 10–180-min lead times using the CSI and HSS metrics over 4–64 mm/h thresholds. Comparisons against optical-flow, deep learning and hybrid baselines show that TUPANN achieves the best or second-best skill in most settings, with pronounced gains at higher thresholds. Training on multiple cities further improves performance, while cross-city experiments show modest degradation and occasional gains for rare heavy-rain regimes. The model produces smooth, interpretable motion fields aligned with numerical optical flow and runs in near real time due to the low latency of GOES‑16. These results indicate that physically aligned learning can provide nowcasts that are skillful, transferable and global.
\end{abstract}

\keywords{precipitation nowcasting, neural networks, physical conditioning, satellite data}

\section{Introduction}

Extreme precipitation events are projected to become more frequent and intense under climate change, increasing the risk of floods and landslides, particularly in vulnerable regions \citep{climate_change}.  Nowcasting—forecasting the atmosphere on time horizons up to 6~h at high spatial resolution—is critical for early warnings and disaster management.  While numerical weather prediction has improved steadily, its finite resolution and latency limit the accuracy of short‑term precipitation forecasts.  Radar‑based nowcasting methods provide detailed observations but often require dense and well‑maintained radar networks that are absent or degraded in much of the world.  For example, Rio de Janeiro experiences recurrent flood‑induced disasters yet lacks reliable radar coverage due to topographic blocking and limited infrastructure.

Recent advances in machine learning have shown that deep networks can outperform traditional numerical models in precipitation nowcasting when trained on high‑resolution radar data.  However, reliance on radar restricts their applicability to radar‑rich regions, leaving large parts of South America, Africa and Asia underserved.  Furthermore, purely data‑driven architectures often struggle with physical interpretability: they may produce realistic‑looking precipitation maps while neglecting physically consistent motion fields, hindering forecasters’ trust and operational uptake.

This paper addresses both accessibility and interpretability by leveraging geostationary satellites, which provide global coverage with near real‑time latency, and by incorporating explicit physical structure into the neural network.  We present TUPANN (Transferable and Universal Physics‑Aligned Nowcasting Network), a model that uses only satellite‑derived precipitation fields and decomposes the forecasting problem into physically motivated submodules.  TUPANN comprises a variational encoder–decoder trained under optical‑flow supervision to recover motion and intensity fields, a lead‑time‑conditioned transformer to evolve latent states, and a differentiable advection operator to reconstruct future frames. A strong physical alignment is done by explicitly penalizing the encoder-decoder output to match results from optical flows algorithms, which numerically infers motion fields. We evaluate TUPANN on data from four climate regimes --- tropical rainforest (Manaus), subtropical highland (La Paz), tropical savanna with coastal influence (Rio de Janeiro) and tropical monsoon (Miami) --- and report critical success indices (CSI) and Heidke skill scores (HSS) across lead times from 10 to 180 minutes and various precipitation thresholds. We benchmark against state‑of‑the‑art baselines, including optical–flow methods (PySTEPS), deep learning models (Earthformer, CasCast), and hybrid approaches (NowcastNet).

Our main contributions are:
\begin{itemize}
    \item We develop a physically aligned satellite‑only nowcasting model that separates motion inference, latent dynamics and advection.  Unlike prior works that learn motion implicitly from final frame loss, our variational encoder–decoder is directly supervised by numerical optical flow, yielding smooth and interpretable motion fields.
    \item We leverage a lead‑time‑conditioned MaxViT transformer to evolve the latent representation and allow long lead‑time prediction with a single network, reducing memory requirements compared with recurrent decoding.
    \item We perform extensive experiments on GOES‑16's Rain Rate Quantitative Precipitation Estimation (RRQPE) and IMERG datasets in up to four cities with different climate regimes, comparing TUPANN with well-established and operational baselines. We demonstrate state‑of‑the‑art CSI and HSS scores at multiple thresholds, analyze the effect of adding a generative adversarial network, and evaluate cross‑city and multi‑city training for transferability.
    \item We discuss operational considerations, including runtime and latency, and outline limitations and future directions for satellite‑based nowcasting.
\end{itemize}

The remainder of this paper is structured as follows.  \Cref{sec:related_work} summarizes related work in numerical, optical‑flow, deep learning and satellite‑only nowcasting. \Cref{sec:data} describes the datasets and study regions. \Cref{sec:methods} details the TUPANN architecture and its components.  \Cref{sec:experiments} presents our experimental design, baselines and results. \Cref{sec:limitations} discusses limitations and future work, and \Cref{sec:conclusion} concludes our work.

\section{Related work}\label{sec:related_work}

\subsection{Numerical nowcasting methods}

Precipitation nowcasting emerged in the late 1980s \citep{https://doi.org/10.1029/RG027i003p00345} and remains a fundamental tool to mitigate the impacts of extreme precipitation events \citep{AN2025126301}. Early approaches relied primarily on Lagrangian extrapolation of radar echoes \citep{ScaleDependence}, while later developments incorporated physical constraints and stochastic perturbations to enable ensemble-based probabilistic forecasts \citep{https://doi.org/10.1002/wrcr.20536}. Among these, PySTEPS \citep{gmd-12-4185-2019} has become a widely adopted open-source Python library providing a reproducible platform for numerical nowcasting. It integrates multiple optical-flow algorithms, including Lucas–Kanade \citep{10.5555/1623264.1623280} and DARTS \citep{darts}, to estimate motion fields and applies the STEPS model \citep{https://doi.org/10.1256/qj.04.100} for probabilistic extrapolation enhanced with downscaled numerical weather prediction input. Despite their interpretability and operational maturity, these numerical schemes typically experience a rapid decline in forecast skill with increasing lead time, and ensemble configurations such as STEPS can incur substantial computational cost.

\subsection{Deep Learning models}

Deep learning (DL) approaches have recently achieved strong performance in precipitation nowcasting, often surpassing traditional numerical methods in both accuracy and scalability while enabling faster inference once trained. Early models include ConvLSTM \citep{10.5555/2969239.2969329} and PredRNN \citep{9749915}, which introduced convolutional and recurrent architectures to capture spatiotemporal dependencies in radar imagery. Transformer-based architectures have since extended this line of work. Earthformer \citep{10.5555/3600270.3602111} adapts the Transformer framework for general Earth-system forecasting through a modified Cuboid Attention mechanism that models three-dimensional spatial interactions. Similarly, Rainformer \citep{9743916} extracts local and global features by combining a window-based multi-head self-attention with a gating mechanism component. These deterministic models, typically trained with pixel-wise L1 or L2 losses, target mean or median intensities and therefore tend to produce overly smooth forecasts.

To address this limitation, generative DL models have been proposed to better reproduce fine-scale variability by learning the underlying data distribution. The first class comprises Generative Adversarial Networks (GANs), including DGMR \citep{Ravuri2021} and NowcastNet \citep{zhang2023}, which have shown competitive performance on radar-based benchmarks. NowcastNet augments the GAN structure with an Evolution Network that estimates motion and intensity fields used to generate intermediate predictions before adversarial refinement. Despite its physics-inspired design, NowcastNet does not explicitly encode physical constraints and inherits known GAN instabilities, including mode collapse and artifact generation \citep{10.1145/3446374}.

More recently, diffusion-based generative models have emerged as stable alternatives to GANs, inspired by advances in computer vision. PreDiff \citep{10.5555/3666122.3669561} employs a Latent Diffusion Model (LDM) with a knowledge-alignment mechanism that enforces domain-specific physical consistency during the sampling process. Evaluated on the SEVIR dataset \citep{10.5555/3495724.3497570}—a combination of GOES-16 satellite imagery and NEXRAD radar data over the United States—PreDiff guides denoising steps toward physically plausible predictions by aligning generated intensities with those estimated from a time-series model applied to context-frame averages. CasCast \citep{10.5555/3692070.3692703} extends this approach through a cascaded LDM framework: it first conditions on deterministic Earthformer predictions and then refines them in latent space to generate high-resolution, small-scale structures. CasCast achieves superior results on SEVIR and other benchmarks, though it remains limited to radar data, short lead times (up to one hour), and lacks explicit physical regularization—sometimes producing noisy outputs and incurring high inference costs typical of diffusion models \citep{salimans2022progressive}.

\subsection{Physically conditioned Deep Learning}

Deep learning models are often regarded as black boxes, offering limited interpretability of the processes guiding their predictions. In geophysical applications, incorporating domain-specific physical knowledge can regularize training and promote physically consistent outputs. Physics-Informed Neural Networks (PINNs) \citep{Raissi2019-cw} exemplify this approach and have inspired numerous extensions and applications \citep{McClenny2023-lu, Kovacs2022-ig, Wang2022-pb, Karniadakis2021-el}. Their core principle is to augment the loss function with a term enforcing that the network outputs satisfy a governing Partial Differential Equation (PDE).

In precipitation nowcasting, several DL models include such physical conditioning. PID-GAN \citep{ca9f2b8763ce44248456c49c7d618ea6} combines a GAN framework with a physics-informed loss derived from the moisture conservation equation, trained on radar data. FourCastNet \citep{10.1145/3592979.3593412} employs an Adaptive Fourier Neural Operator architecture \citep{Guibas2021AdaptiveFN}, a physics-inspired design widely used for PDE solutions, and applies it to global-scale forecasting with ERA5 reanalysis inputs. The previously discussed PreDiff and NowcastNet also incorporate elements of physical conditioning: NowcastNet’s design draws inspiration from the continuity equation without numerical constraints, whereas PreDiff explicitly penalizes deviations from numerical-model intensities. The conditioning proposed in this work builds on both ideas by enforcing the continuity equation through an explicit loss between predicted physical terms and those derived from a numerical optical-flow method, applied to a specific module of the architecture. As shown later, this formulation achieves competitive predictive skill while improving interpretability and physical plausibility.

\subsection{Use of satellite data}

The use of satellite observations for precipitation nowcasting remains relatively limited. Several studies have explored this direction. \citet{2025MAP...137....3S} evaluated the use of PySTEPS with satellite imagery under different optical flow methods. \citet{10.1145/3292500.3330762} employed a variant of the U-Net architecture \citep{10.1007/978-3-319-24574-4_28} to predict precipitation up to two hours ahead over Russia using EUMETSAT data. \citet{GlobalPrecipitationNowcastingofIntegratedMultisatellitERetrievalsforGPMAUNetConvolutionalLSTMArchitecture} proposed a hybrid U-Net–ConvLSTM model evaluated on IMERG and Global Forecast System (GFS) data, using GFS as ground truth but without comparison against baseline models. More recently, \citet{10.1609/aaai.v39i27.35049} introduced a two-phase neural network that first predicts future satellite imagery using a video prediction model and then performs image-to-image translation to obtain radar reflectivity. Their approach leverages the Sat2Rdr dataset, derived from the Korean GK2A geostationary satellite and ten ground-based radar stations. While effective for light precipitation, reported results are limited to low critical success index (CSI) thresholds (below 8 mm/h), leaving high-intensity events mostly unassessed. \cite{agrawal2025} also leverage geostationary satellite mosaics, providing global skillful forecasts up to 12 hours into the future. The model uses an encoder-decoder architecture with multiple high-dimensional inputs, including numerical weather prediction (NWP), leading to an intensive hardware use of more than 500 TPU cores during training.

Other studies, including \citet{10472033}, \citet{10662919}, and \citet{andrychowicz2023deeplearningdayforecasts}, integrate both satellite and radar data as inputs, which limits their applicability in radar-sparse regions. In contrast, the present work relies exclusively on satellite imagery, enabling global scalability. Comparisons are conducted against established baselines and evaluated across a range of precipitation intensities, including high and extreme-rate events.

\section{Data and study regions}\label{sec:data}

We train and evaluate TUPANN using precipitation data from satellite products. For each dataset we identify rain events and split them into training, validation and test sets as described below.

\subsection{GOES‑16 RRQPE}
\begin{figure}[ht]
    \centering
    \includegraphics{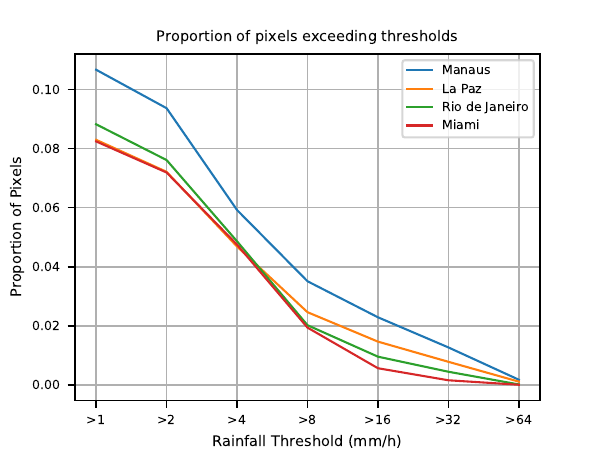}
    \caption{Proportion of observations above different precipitation thresholds in the GOES‑16 RRQPE dataset for each of the four study regions.  Manaus exhibits the highest frequency of heavy rainfall across thresholds, while Rio de Janeiro and La Paz show intermediate levels}
    \label{fig:prop_threshold_goes}
\end{figure}

\begin{figure}[ht]
  \centering
  \includegraphics[scale=0.8]{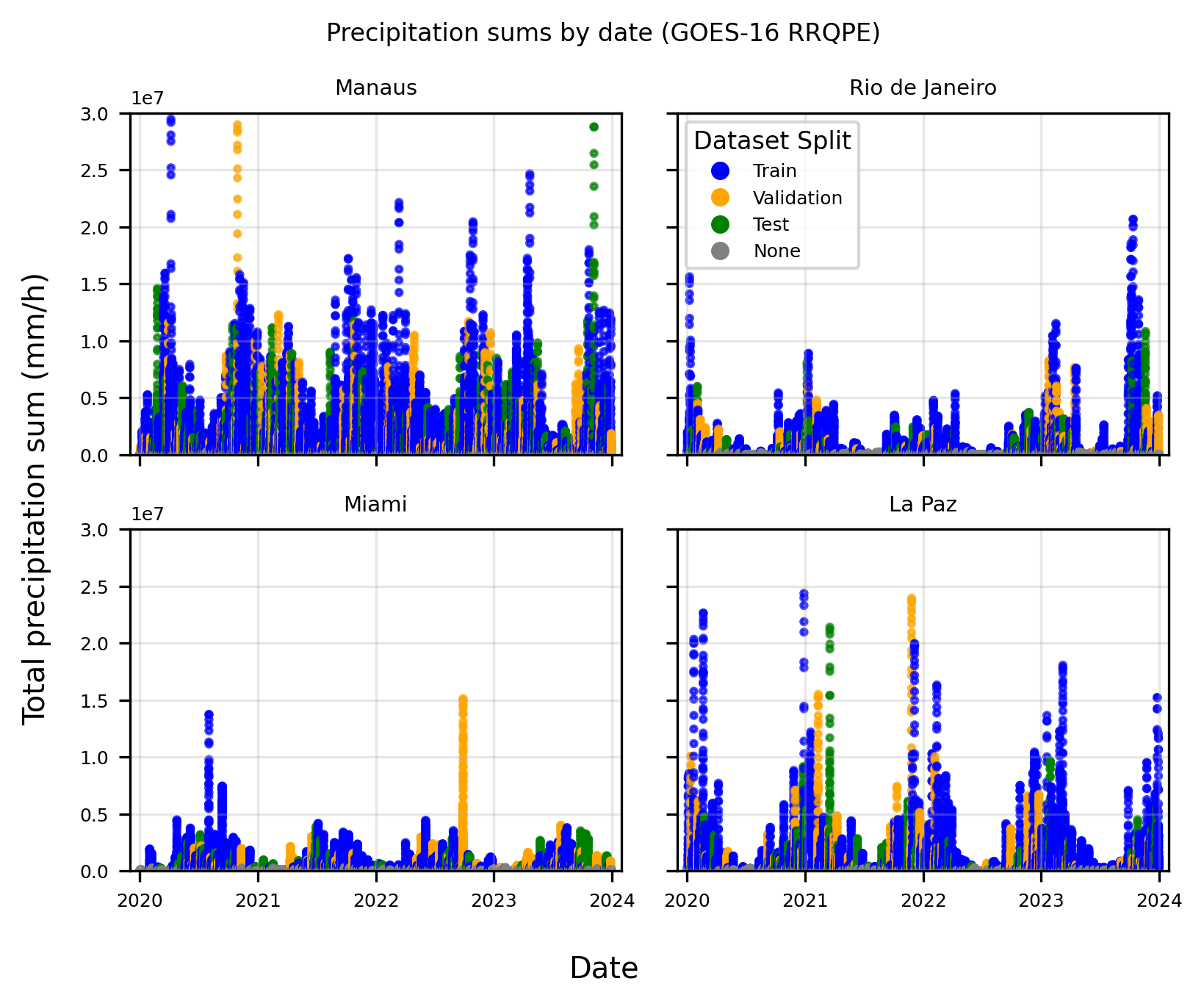}
  \caption{Accumulated precipitation in GOES‑16 RRQPE from January~2020 to December~2023 over each study region.  Shaded areas denote training, validation and test splits.  Seasonal variability differs markedly between regions, with pronounced dry and wet seasons in La Paz and Rio de Janeiro}
  \label{fig:total_sums}
\end{figure}

The primary data source is the GOES‑R Advanced Baseline Imager Rain Rate Quantitative Precipitation Estimation (RRQPE) \citep{RRQPE}. RRQPE provides precipitation estimates over the Americas every 10~min at \(2\,\text{km}\) spatial resolution with a latency of approximately 5~min, enabling real‑time nowcasting. This product is highly correlated with rain-related bands and has been validated against ground radars and the GPM CORRA dataset \citep{agrawal2025}, highlighting the value of predicting geostationary satellite observations. We use RRQPE from January~2020 to December~2023. Rain events are defined as contiguous periods when the precipitation rate exceeds a chosen threshold (see \Cref{sec:rain_events} for details); we sample events uniformly at random and allocate 70\% to training, 15\% to validation and 15\% to testing. \Cref{fig:prop_threshold_goes} shows the proportion of observations above various thresholds for four different cities, while 
\Cref{fig:total_sums} displays the accumulated precipitation and dataset splits.

\subsection{IMERG}
\begin{figure}[htpb]
    \centering
    \begin{subfigure}{.5\textwidth}
      \centering
      \includegraphics{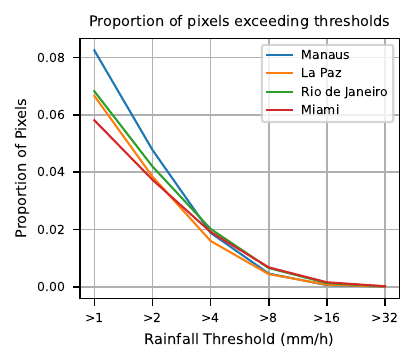}
      \caption{Proportion of observations above different precipitation thresholds in the IMERG dataset for each study region.  Unlike RRQPE, Manaus only dominates the lowest thresholds.}
    \end{subfigure}%
    \begin{subfigure}{.5\textwidth}
      \centering
      \includegraphics{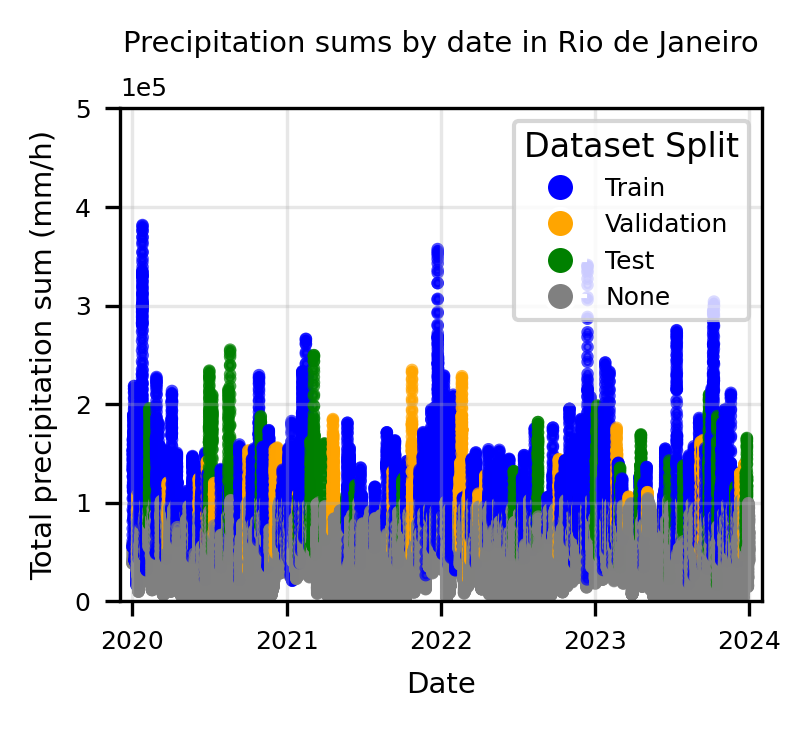}
      \caption{Estimated precipitation from IMERG (2020–2023) over Rio de Janeiro. The data splits are color-coded; low‑level precipitation events are excluded from training and evaluation.}
    \end{subfigure}
    \caption{Statistics of IMERG data, highlighting differences with respect to the RRQPE dataset}
    \label{fig:prop_threshold_imerg}
\end{figure}

To test generalization across data sources we also use the Integrated Multi‑satellitE Retrievals for GPM (IMERG) Final Run product \citep{IMERG}.  IMERG provides precipitation estimates every 30~min at \(10\,\text{km}\) resolution and is widely used in remote sensing research.  Its latency is about 3.5~months (the Early Run version has 4~h latency), which precludes real‑time use but offers an independent validation dataset.  We extract IMERG data from January~2020 to December~2023, split it using the same event‑based procedure, and consider only the Rio de Janeiro region.  \Cref{fig:prop_threshold_imerg} compares the proportion of heavy‑rain observations and accumulated precipitation for IMERG.

\subsection{Study regions}
To evaluate model performance across different climates we select four  \(\textrm{512 km} \times 512\textrm{ km}\) subregions of the GOES‑16 domain centered on Rio de Janeiro (Brazil), La Paz (Bolivia), Manaus (Brazil) and Miami (USA). These regions span coastal, high‑altitude, rainforest and subtropical environments. For IMERG we consider only a \(2560\textrm{ km} \times 2560\textrm{ km}\) area surrounding Rio de Janeiro. The dominant precipitation processes include orographic and mesoscale convection in Rio \citep{silva2022rio}, high‑altitude convective storms in La Paz \citep{Garreaud2001}, monsoon‑driven convection in Manaus \citep{OliveiraEtAl2016} and sea‑breeze thunderstorms in Miami \citep{Burpee1979}.  \Cref{fig:total_sums} illustrates the seasonal cycle and dataset splits for GOES‑16 across regions.

\section{Methods}\label{sec:methods}

\begin{figure}[ht]
    \centering
    \includegraphics[scale=0.25]{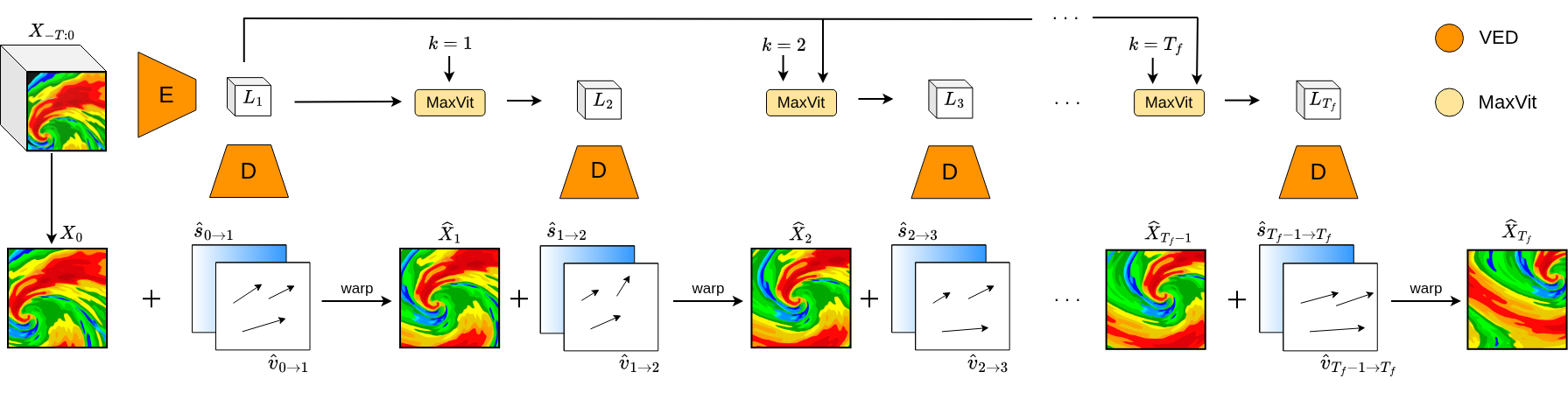}
    \caption{TUPANN architecture. The VED and MaxViT modules displayed are learned; motion fields and the final predictions are extrapolated through a warp function}
    \label{fig:tupann_architecture}
\end{figure}

The proposed model, TUPANN, forecasts precipitation fields from a sequence of past satellite images \(X_{-T:0}\in \R^{(T+1)\times n\times n}\), where $n$ denotes the spatial resolution (i.e., number of pixels per dimension) and $T+1$ is the number of past observations.  It produces predicted fields \(\widehat{X}_{1:T_f}\in \R^{T_f\times n\times n}\), where $T_f$ is the forecast horizon. The architecture comprises two modules: (i) a variational encoder–decoder (VED) that learns a latent representation $L_k$, $k=1, \ldots, T_f$ of the evolution of the precipitation fields under optical flow supervision; (ii) a visual transformer (MaxViT) that evolves the latent representation $L_k$ so that the resulting application of a differentiable advection operator (warp) on the decoded motion and intensity fields, $\widehat{s}_{k-1\rightarrow k}$ and $\widehat{v}_{k-1\rightarrow k}$,  closely match the ground truth frame. See \Cref{fig:tupann_architecture}. The training procedure is sequential: the VED module is initially trained to infer the first set of motion and intensity fields, then its weights are fixed and used for the training of the MaxVit module. Details on the VED and MaxViT training are discussed in Sections \ref{sec:ved} and \ref{sec:MaxViT}, respectively.

Even though MaxViT offers linear complexity in the image size, the choice of such encoder-decoder architecture is guided by the idea that, apart from compressing the images, the VED is also responsible for learning the dynamics of a single step evolution. MaxViT, on the other hand, learns to extrapolate the dynamics to further lead times.

\subsection{Variational encoder–decoder}\label{sec:ved}

\begin{figure}[ht]
    \centering
    \includegraphics[scale=0.8]{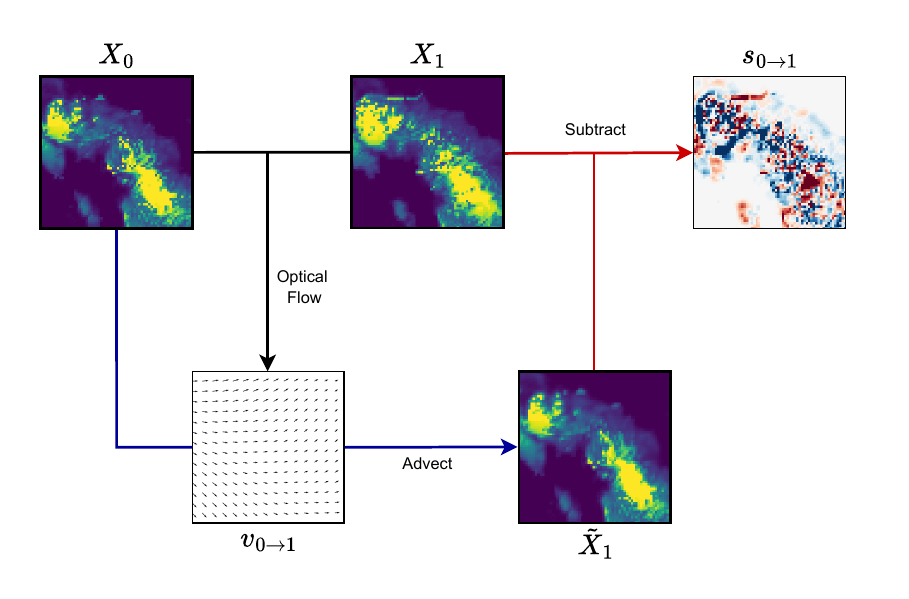}
    \caption{Ground truth motion fields are obtained using an optical flow method from a pair of past and future images. The past image is advected to obtain an intermediate one, $\tilde{X}_1$. Finally, ground truth intensity is the subtraction of $\tilde{X}_1$ from the future image}
    \label{fig:ground_truth_fields}
\end{figure}

We use a variational encoder–decoder to learn an efficient representation of the precipitation evolution. Instead of reconstructing the input images as in classical variational autoencoders, our VED outputs motion fields \(\widehat{v}_{0\rightarrow 1}\in \R^{2\times n\times n}\) and intensity corrections \(\widehat{s}_{0\rightarrow 1}\in \R^{n\times n}\) given the past sequence \(X_{-T:0}\). To enforce physically plausible motion we compute the ground truth motion fields \(v_{0\rightarrow1}\) applying an optical flow algorithm to \( X_{-\tilde{T}+2:1} \), where \( \tilde{T} \) is the context length provided to the optical flow algorithm. After that, the ground truth intensity correction \(s_{0\rightarrow1}\) is obtained by subtracting the advected frame $\tilde{X}_1$, obtained using $v_{0\rightarrow1}$, from the true frame \(X_1\) (see \Cref{fig:ground_truth_fields}). Thus, the estimated fields \(\widehat{v}_{0\rightarrow 1}\) and \(\widehat{s}_{0\rightarrow 1}\) can be used to extrapolate the last observed frame \(X_0\) to an estimate \(\widehat{X}_1\) of the next frame via an advection operator (see \Cref{sec:warp}). 

\subsubsection{Target loss}

To supervise the predicted \(\widehat{v}_{0\rightarrow1}\), \(\ell_1\) and cosine‑similarity losses are used with respect to  \(v_{0\rightarrow1}\). The intensity discrepancies between $s_{0\rightarrow1}$ and $\widehat{s}_{0\to1}$ are penalized via $\ell_1$ loss. Finally, a Kullback-Leibler divergence term is added to ensure the regularity of the learned latent space. The VED loss is
\begin{equation}
      \begin{aligned}
        \text{Loss}_{\textrm{VED}}(\widehat{s}_{0\rightarrow1}, \widehat{v}_{0\rightarrow1}, s_{0\rightarrow1}, v_{0\rightarrow1})  := & \; \lambda_{\textrm{int}} \: \ell_1(s_{0\rightarrow1}, \widehat{s}_{0\rightarrow1}) + \lambda_{\textrm{motion}} \: \ell_1(v_{0\rightarrow1}, \widehat{v}_{0\rightarrow1}) \\
        & + \lambda_{\textrm{cos}} \: \text{CosSimilarity}(v_{0\rightarrow1}, \widehat{v}_{0\rightarrow1}) + \lambda_{\textrm{KL}} \:KL(p,\widehat{p}_\theta).
    \end{aligned}
    \label{loss_autoenc}
    \end{equation}
Here, \(\widehat{p}_\theta\) is the latent distribution inferred by the encoder, \(p\) is a standard normal distribution and \(\lambda_{\mathrm{int}},\lambda_{\mathrm{motion}},\lambda_{\mathrm{cos}},\lambda_{\mathrm{KL}}\) are hyperparameters tuned on the validation set.  This loss encourages accurate motion fields, intensity corrections and latent regularity.

\subsubsection{Optical flow}

An optical flow algorithm is able to infer motion fields between two images, and is thus essential to obtain the derived ground-truth motion and intensity fields (e.g., $v_{0\to1}, s_{0\to1}$) used in \eqref{loss_autoenc}. We consider two options: Lucas–Kanade (LK), which solves a local least‑squares problem under the assumption of small displacements, and DARTS, a spectral method tailored to radar imagery that solves the optical‑flow equation in Fourier space. For TUPANN, we have taken the choice of optical flow method as a hyperparameter; below we use DARTS for GOES-16 and LK for IMERG results.

\subsection{MaxViT}\label{sec:MaxViT}

Given the latent representation \(L_1\) from the VED, a visual transformer evolves the latent state forward in time. We adopt MaxViT \citep{Tu2022Nov}, which combines local and grid attention to efficiently capture global context while avoiding quadratic attention cost. 

\subsubsection{Lead time conditioning}

To predict the latent state at lead time \(k\), we condition the transformer on \(k\) via one‑hot encoding and linear embedding, yielding \(L_k = \mathrm{VT}(L_1,k)\), $k=2, \ldots, T_f$, where where $\mathrm{VT}(\cdot)$ represents the MaxViT model. Unlike recurrent decoding, this conditioning enables the same transformer to produce all lead times while reducing memory overhead \citep{andrychowicz2023deeplearningdayforecasts}.  Applying the VED decoder to \(L_k\) yields motion and intensity fields \(\big(\widehat{v}_{k-1\rightarrow k},\widehat{s}_{k-1\rightarrow k}\big)=D(L_k)\), where $D(\cdot)$ is the decoder module of the VED. Thus, all the necessary elements to predict the sequence recursively are obtained.

\subsubsection{Warp function}\label{sec:warp}

Following NowcastNet \citep{zhang2023}, we implement a fixed differentiable advection operator that can reconstruct future precipitation frames using the predicted motion and intensity fields. Thus, given a frame \(\widehat{X}_{k-1}\), motion field \(\widehat{v}_{k-1 \to k}\) and intensity field \(\widehat{s}_{k-1 \to k}\), the extrapolated frame $\widehat{X}_k$ is
\begin{equation}
 \widehat{X}_k = \text{warp} \left( \widehat{v}_{k-1\rightarrow k}, \widehat{s}_{k-1\rightarrow k},  \widehat{X}_{k-1}\right).
\label{warp_func}
\end{equation}

\subsubsection{Target loss}

We compute the target loss in the original image space. We assume that $\widehat{X}_{k-1} = X_{k-1}$ in equation (\ref{warp_func}) to avoid a costly recursive loss and calculate the $\ell_1$ loss between the warped frame $\widehat{X}_{k}$ and the observed frame $X_k$. Thus,
\begin{equation}
 \text{Loss}_{\textrm{MaxViT}} = \ell_1 \left(\text{warp} \left( \widehat{v}_{k-1\rightarrow k}, \widehat{s}_{k-1\rightarrow k}, X_{k-1} \right), X_k \right).
\label{MaxViT_loss}
\end{equation}
The gradients of this loss will flow through the VED decoder module and the MaxViT transformer. The VED is pre-trained separately, thus optimizing this loss only affects the MaxViT modules.

\section{Experiments and results}\label{sec:experiments}

We compare TUPANN against four nowcasting benchmarks using several evaluation metrics and across regions with different climates. We also run several ablation experiments.

\subsection{Evaluation framework}

We evaluate TUPANN and baselines using the critical success index (CSI) and Heidke skill score (HSS).  
Both metrics depend on a choice of threshold $t$, so that pixel values above the threshold are assigned as positive and, otherwise, negative. CSI and HSS are given by
\begin{equation*}
    \textrm{CSI}_t = \frac{\textrm{TP}_{t}}{\textrm{TP}_t + \textrm{FN}_t +\textrm{FP}_t}, \qquad \textrm{HSS}_t = \frac{2\left(\textrm{TP}_t \times \textrm{TN}_t - \textrm{FN}_t \times \textrm{FP}_t\right)}{\left(\textrm{TP}_t + \textrm{FN}_t\right)\left(\textrm{TN}_t + \textrm{FN}_t\right) - \left(\textrm{TP}_t + \textrm{FP}_t\right)\left(\textrm{TN}_t + \textrm{FP}_t\right)},
\end{equation*}
where $\textrm{TP}$ stands for true positive, \( \text{TN} \) for true negatives, $\textrm{FN}$ for false negatives and $\textrm{FP}$ for false positives. Thus, CSI quantifies the overlap between forecasted and observed precipitation, ignoring true negatives (i.e., disregarding correct predictions of no precipitation) while HSS compares the forecast against random chance.

We report CSI and HSS at thresholds of 4, 8, 16, 32 and 64~mm/h and compute both pixel‑wise scores (POOL1) and max-pooled scores over \(4\times4\) blocks (POOL4).  For aggregated metrics we denote CSI–M and HSS–M as the mean across all thresholds. CSI values are reported in \Cref{sec:results}, and HSS ones are included in \Cref{appendixA}.

\subsection{Baseline models and tuning}\label{subsec:baselines}
In our experiments, TUPANN is compared with four baselines from different nowcasting paradigms:
\begin{itemize}
    \item PySTEPS (LK) and PySTEPS (DARTS) \citep{gmd-12-4185-2019}: 
    optical-flow baselines that estimate a motion field (LK: local Lucas–Kanade; DARTS: DFT-based spectral) from recent frames and then semi-Lagrangianly advect the precipitation field forward. As purely physical extrapolation methods, they are strong for very short lead times;

    \item Earthformer \citep{10.5555/3600270.3602111}: 
    a space-time Transformer for Earth-system data that uses Cuboid Attention (local block attention with global tokens) in a hierarchical encoder–decoder to predict future frames;
    
    \item NowcastNet \citep{zhang2023}: a hybrid model combining a U-Net–based learnable semi-Lagrangian advection (Evolution Network) with a physics-conditioned generative network trained with a temporal discriminator to inject high-resolution convective structure;
    
    \item CasCast \citep{cascast}: a cascaded scheme that first uses a deterministic predictor (e.g., Earthformer) to capture mesoscale evolution, then conditions a latent-space diffusion transformer on that coarse forecast to generate small-scale features and improve extreme-precipitation skill.
\end{itemize}

For TUPANN and Earthformer, we tune learning rate, dropout rate and loss weights on the validation set by maximizing mean CSI in the city of Rio de Janeiro, and use these for the other cities. Hyperparameters for NowcastNet and CasCast are mostly those presented in their original paper (see \Cref{appendixA}). The optimizer for all models is Adam. After selecting the best values, we retrain on the combined training and validation data and evaluate on a held-out test set. Training uses a single NVIDIA A100 GPU, and inference typically takes under two seconds per forecast (for all 18 lead times).

\subsection{GOES‑16 results} \label{sec:results}
\Cref{tab:csi_final} presents CSI scores across the four study regions and thresholds. TUPANN consistently ranks first or second for each metric.  In Rio de Janeiro, it achieves the highest CSI at all thresholds; NowcastNet is the closest competitor, followed by Earthformer. In Miami, TUPANN again dominates most metrics, but now CasCast performs well.  In Manaus and La Paz, Earthformer and NowcastNet obtain slightly better CSI for low thresholds, but TUPANN leads for higher thresholds and is therefore better at forecasting extreme rainfall.  The 64~mm/h CSI values are small across models, reflecting the rarity of such intense events, yet TUPANN’s scores remain the highest. Overall, the table highlights TUPANN's performance across different climate regimes and precipitation thresholds. Similar results are true for HSS (see \Cref{tab:hss_final}), although by that metric Earthformer is much more competitive. 

\begin{table*}[ht]
\centering
\caption{Aggregated CSI metrics for GOES‑16 data across cities.  \textbf{Bold} values denote the best, \underline{underlined} values the second best.  TUPANN obtains state‑of‑the‑art performance at most thresholds and regions, particularly for high rain‑rate events.}
\renewcommand{\arraystretch}{1.3}
\resizebox{1.0\columnwidth}{!}{
\begin{tabular}{l|cc|cc|cc|cc|cc|cc}
\toprule
\multirow{2}{*}{Model}
 & \multicolumn{2}{c|}{$\textrm{CSI-M}\uparrow$}
 & \multicolumn{2}{c|}{$\textrm{CSI}_{4}\uparrow$}
 & \multicolumn{2}{c|}{$\textrm{CSI}_{8}\uparrow$}
 & \multicolumn{2}{c|}{$\textrm{CSI}_{16}\uparrow$}
 & \multicolumn{2}{c|}{$\textrm{CSI}_{32}\uparrow$}
 & \multicolumn{2}{c}{$\textrm{CSI}_{64}\uparrow$} \\
\cline{2-13}
 & POOL1 & POOL4 & POOL1 & POOL4 & POOL1 & POOL4 & POOL1 & POOL4 & POOL1 & POOL4 & POOL1 & POOL4 \\
\hline

& \multicolumn{12}{|c}{\textbf{Rio de Janeiro}\rule{0pt}{2.5ex}}\\
\hline
Earthformer & 0.237 & 0.222 & \underline{0.326} & 0.320 & \underline{0.287} & 0.236 & \underline{0.326} & \underline{0.312} & 0.238 & 0.237 & 0.009 & 0.006 \\
NowcastNet & \underline{0.244} & \underline{0.269} & 0.313 & \underline{0.374} & 0.282 & \textbf{0.293} & 0.318 & 0.325 & \underline{0.247} & \underline{0.278} & \underline{0.059} & \underline{0.074} \\
PySTEPS (LK) & 0.165 & 0.169 & 0.242 & 0.262 & 0.212 & 0.195 & 0.226 & 0.226 & 0.142 & 0.156 & 0.005 & 0.008 \\
PySTEPS (DARTS) & 0.166 & 0.166 & 0.231 & 0.243 & 0.216 & 0.191 & 0.229 & 0.228 & 0.140 & 0.152 & 0.013 & 0.015 \\
CasCast & 0.170 & 0.187 & 0.308 & 0.343 & 0.205 & 0.249 & 0.164 & 0.162 & 0.159 & 0.156 & 0.016 & 0.027 \\
\hline
\textbf{TUPANN (ours)} & \textbf{0.259} & \textbf{0.277} & \textbf{0.330} & \textbf{0.384} & \textbf{0.289} & \underline{0.289} & \textbf{0.330} & \textbf{0.336} & \textbf{0.274} & \textbf{0.287} & \textbf{0.072} & \textbf{0.090} \\
\hline

& \multicolumn{12}{|c}{\textbf{Miami}\rule{0pt}{2.5ex}}\\
\hline
Earthformer & 0.141 & 0.126 & \textbf{0.274} & 0.270 & 0.180 & 0.160 & \underline{0.154} & 0.122 & 0.097 & 0.078 & 0.000 & 0.000 \\
NowcastNet & 0.137 & 0.160 & 0.248 & \underline{0.299} & 0.170 & 0.207 & 0.128 & 0.139 & 0.097 & 0.106 & 0.040 & \underline{0.047} \\
PySTEPS (LK) & 0.113 & 0.116 & 0.188 & 0.202 & 0.133 & 0.136 & 0.120 & 0.111 & 0.079 & 0.079 & \underline{0.045} & 0.053 \\
PySTEPS (DARTS) & 0.112 & 0.113 & 0.189 & 0.202 & 0.135 & 0.138 & 0.118 & 0.107 & 0.073 & 0.071 & 0.044 & 0.046 \\
CasCast & \underline{0.146} & \underline{0.170} & 0.258 & 0.298 & \underline{0.188} & \textbf{0.229} & 0.144 & \underline{0.167} & \underline{0.117} & \underline{0.135} & 0.020 & 0.028 \\
\hline
\textbf{TUPANN (ours)} & \textbf{0.169} & \textbf{0.187} & \underline{0.267} & \textbf{0.312} & \textbf{0.189} & \underline{0.211} & \textbf{0.177} & \textbf{0.177} & \textbf{0.135} & \textbf{0.141} & \textbf{0.079} & \textbf{0.094} \\
\hline

& \multicolumn{12}{|c}{\textbf{Manaus}\rule{0pt}{2.5ex}}\\
\hline
Earthformer & \underline{0.276} & 0.256 & \textbf{0.355} & 0.341 & \textbf{0.323} & 0.297 & \textbf{0.316} & 0.292 & \underline{0.265} & 0.245 & 0.124 & 0.104 \\
NowcastNet & 0.253 & 0.278 & 0.323 & 0.366 & 0.296 & \underline{0.324} & 0.283 & 0.303 & 0.233 & 0.258 & 0.130 & 0.137 \\
PySTEPS (LK) & 0.200 & 0.196 & 0.258 & 0.266 & 0.237 & 0.233 & 0.218 & 0.212 & 0.160 & 0.156 & 0.125 & 0.112 \\
PySTEPS (DARTS) & 0.197 & 0.194 & 0.259 & 0.268 & 0.239 & 0.235 & 0.219 & 0.213 & 0.158 & 0.154 & 0.109 & 0.099 \\
CasCast & 0.265 & \underline{0.286} & \underline{0.344} & \textbf{0.377} & 0.303 & \textbf{0.333} & 0.295 & \underline{0.307} & 0.260 & \underline{0.269} & \underline{0.126} & \underline{0.141} \\
\hline
\textbf{TUPANN (ours)} & \textbf{0.290} & \textbf{0.293} & 0.339 & \underline{0.367} & \underline{0.316} & 0.321 & \underline{0.315} & \textbf{0.312} & \textbf{0.278} & \textbf{0.274} & \textbf{0.200} & \textbf{0.193} \\
\hline

& \multicolumn{12}{|c}{\textbf{La Paz}\rule{0pt}{2.5ex}}\\
\hline
Earthformer & \underline{0.303} & 0.270 & \textbf{0.337} & 0.312 & \textbf{0.329} & 0.281 & \textbf{0.359} & \underline{0.319} & \underline{0.323} & 0.291 & 0.167 & 0.146 \\
NowcastNet & 0.291 & \underline{0.301} & 0.330 & \textbf{0.376} & 0.303 & \underline{0.321} & 0.321 & 0.315 & 0.300 & \underline{0.296} & \underline{0.202} & \underline{0.197} \\
PySTEPS (LK) & 0.212 & 0.208 & 0.248 & 0.250 & 0.243 & 0.230 & 0.247 & 0.237 & 0.197 & 0.195 & 0.126 & 0.131 \\
PySTEPS (DARTS) & 0.225 & 0.218 & 0.263 & 0.262 & 0.262 & 0.244 & 0.264 & 0.250 & 0.206 & 0.201 & 0.127 & 0.132 \\
CasCast & 0.228 & 0.235 & 0.309 & 0.337 & 0.251 & 0.270 & 0.245 & 0.237 & 0.232 & 0.222 & 0.101 & 0.111 \\
\hline
\textbf{TUPANN (ours)} & \textbf{0.314} & \textbf{0.317} & \underline{0.336} & \underline{0.363} & \underline{0.327} & \textbf{0.323} & \underline{0.350} & \textbf{0.340} & \textbf{0.327} & \textbf{0.322} & \textbf{0.232} & \textbf{0.239} \\
\bottomrule
\end{tabular}}
\label{tab:csi_final}
\end{table*}

\begin{figure}[ht]
    \centering
    \includegraphics[width=0.8\textwidth]{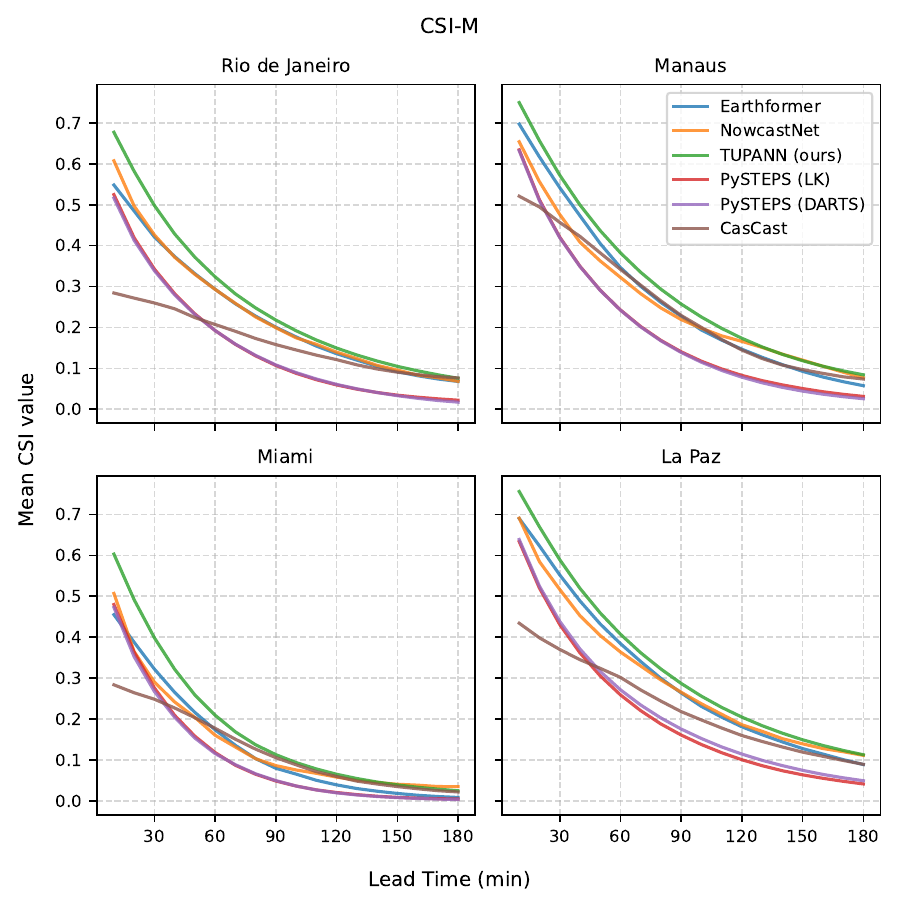}
    \caption{Mean CSI (CSI–M) versus lead time for the four study regions using GOES‑16 data.  TUPANN consistently outperforms baselines across lead times}
    \label{fig:avg_csi_goes16}
\end{figure}

The graphs in \Cref{fig:avg_csi_goes16} show mean CSI (averaged across thresholds) versus lead time. TUPANN maintains the highest or second‑highest skill across all lead times; the advantage over NowcastNet grows for early lead times, reflecting the benefit of explicit motion supervision and the efficiency of lead‑time conditioning (see also \Cref{fig:fields}).

Beyond aggregated metrics, \Cref{fig:prediction_example_big} illustrates a TUPANN prediction for a rain event in Manaus, compared with NowcastNet, CasCast and Earthformer.  Generative models such as NowcastNet and CasCast produce detailed textures but may introduce artifacts, whereas TUPANN and Earthformer yield smoother predictions.  Despite the blurred appearance, TUPANN captures the timing and location of heavy rain more accurately, leading to higher CSI values.

\subsection{Ablation results}

We will study how different experiments affect TUPANN: investigating its motion fields against another physics-DL hybrid model, adding a GAN head, evaluating TUPANN's cross-city generalization and training the model jointly on all cities.

\subsubsection{Interpretability and motion fields}

\begin{figure}[!ht]
    \centering
    \includegraphics[width=\textwidth]{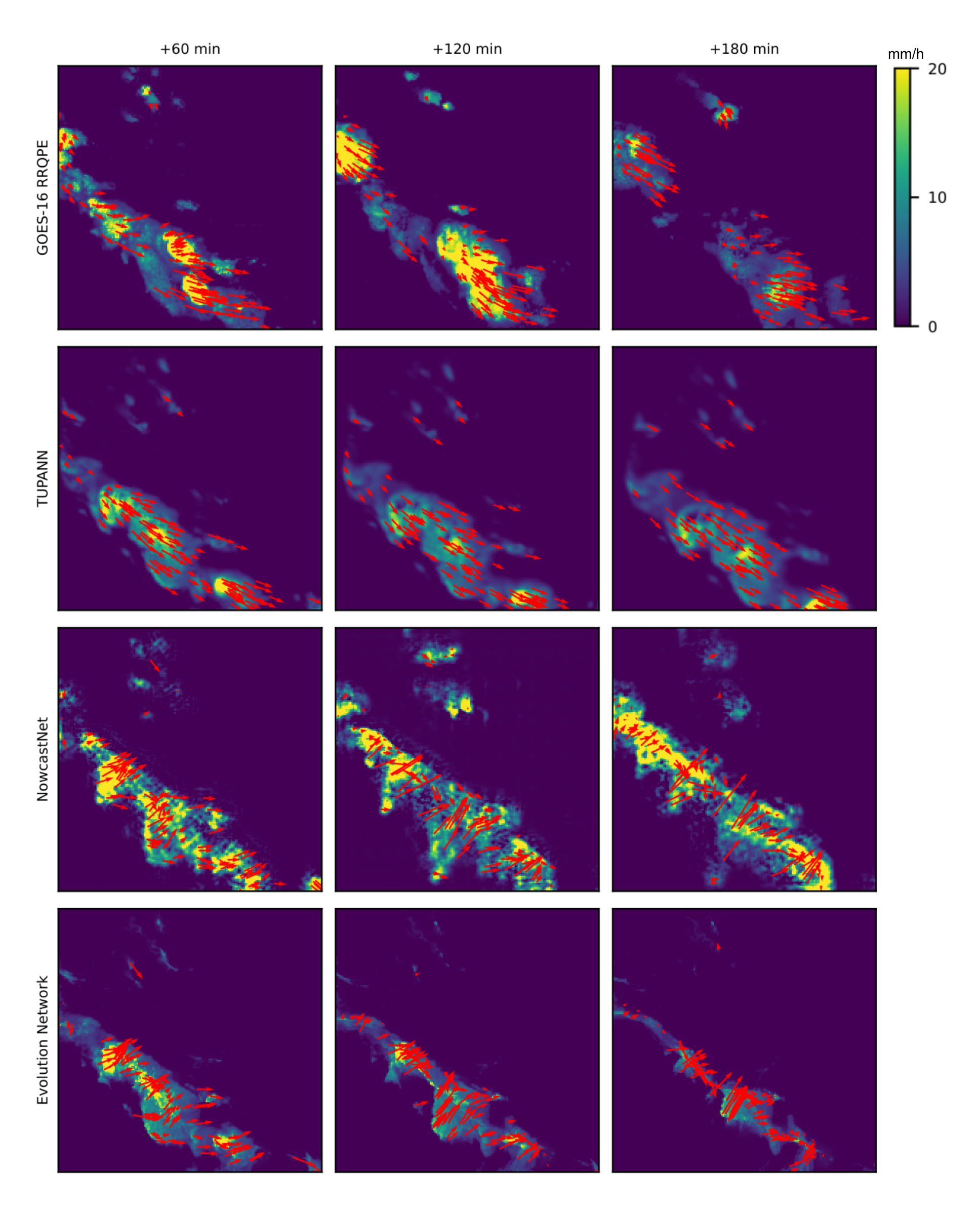}
    \caption{Comparison of motion fields.  Top row: ground‑truth DARTS motion fields for future frames.  Second row: motion fields from TUPANN.  Subsequent rows: motion fields estimated by NowcastNet/Evolution Network.  TUPANN yields smoother fields that align with physical intuition}
    \label{fig:fields}
\end{figure}

TUPANN's interpretability stems from its explicitly learned motion fields.  \Cref{fig:fields} compares motion fields predicted by TUPANN and NowcastNet (which relies on an Evolution Network submodule for its motion fields).  The TUPANN fields are smooth and closely resemble the numerical optical flow computed by DARTS, whereas the baselines’ fields exhibit unrealistic patterns. This underscores the benefit of supervising motion fields directly.

\subsubsection{GAN-TUPANN}

Generative adversarial networks can improve visual realism at the cost of evaluation metrics. To study its impact on TUPANN predictions, we evaluate the variant GAN‑TUPANN, which adds a GAN head to TUPANN outputs. 

\begin{figure}[!ht]
    \centering
    \includegraphics[scale=0.5]{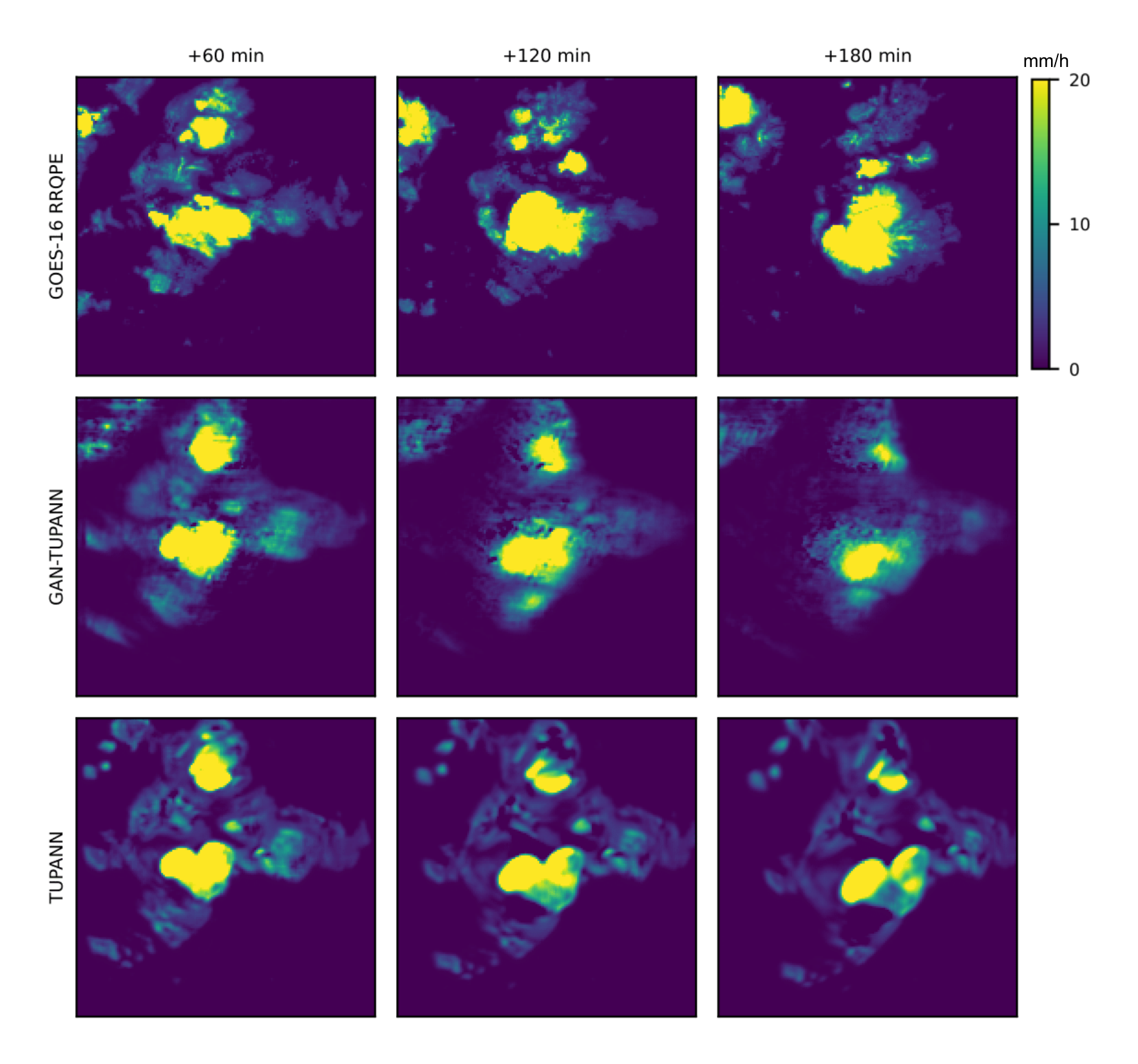}
    \caption{Visual comparison of TUPANN and GAN‑TUPANN for the same Manaus event as in \Cref{fig:prediction_example_big}.  GAN‑TUPANN reduces blur but yields mixed changes in CSI (\Cref{tab:csi_gan_TUPANN})}
    \label{fig:gan}
\end{figure}

\Cref{fig:gan} shows that GAN‑TUPANN produces significantly sharper images. Still, \Cref{tab:csi_gan_TUPANN} shows this does not always lead to improvements in CSI scores. While in Rio de Janeiro GAN‑TUPANN increases low‑threshold CSI, in Miami the gains are either non-existent or negative. Given the computational overhead and mixed impact on metrics, there is no clear advantage in including this module unless visual fidelity is paramount.  

\begin{table*}[ht]
\centering
\caption{CSI metrics comparing TUPANN and its GAN variant (GAN‑TUPANN) on GOES‑16 data.  Bold values denote the best, underlined values the second best.  GAN‑TUPANN improves low‑threshold performance in Rio de Janeiro but degrades or yields marginal improvements in other cities.}
\renewcommand{\arraystretch}{1.3}
\resizebox{\textwidth}{!}{
\begin{tabular}{l|cc|cc|cc|cc|cc|cc}
\toprule
\multirow{2}{*}{Model}
 & \multicolumn{2}{c|}{$\textrm{CSI-M}\uparrow$}
 & \multicolumn{2}{c|}{$\textrm{CSI}_{4}\uparrow$}
 & \multicolumn{2}{c|}{$\textrm{CSI}_{8}\uparrow$}
 & \multicolumn{2}{c|}{$\textrm{CSI}_{16}\uparrow$}
 & \multicolumn{2}{c|}{$\textrm{CSI}_{32}\uparrow$}
 & \multicolumn{2}{c}{$\textrm{CSI}_{64}\uparrow$} \\
\cline{2-13}
 & POOL1 & POOL4 & POOL1 & POOL4 & POOL1 & POOL4 & POOL1 & POOL4 & POOL1 & POOL4 & POOL1 & POOL4 \\
\hline

& \multicolumn{12}{|c}{\textbf{Rio de Janeiro}\rule{0pt}{2.5ex}}\\
\hline
NowcastNet & 0.244 & 0.269 & 0.313 & 0.374 & 0.282 & \underline{0.293} & 0.318 & 0.325 & 0.247 & 0.278 & 0.059 & 0.074 \\
\hline
\textbf{TUPANN} & \underline{0.259} & \underline{0.277} & \underline{0.330} & \underline{0.384} & \underline{0.289} & 0.289 & \underline{0.330} & \underline{0.336} & \textbf{0.274} & \textbf{0.287} & \textbf{0.072} & \textbf{0.090} \\
\textbf{GAN‑TUPANN} & \textbf{0.265} & \textbf{0.290} & \textbf{0.350} & \textbf{0.413} & \textbf{0.296} & \textbf{0.319} & \textbf{0.342} & \textbf{0.348} & \underline{0.270} & \underline{0.286} & \underline{0.070} & \underline{0.086} \\
\hline

& \multicolumn{12}{|c}{\textbf{Miami}\rule{0pt}{2.5ex}}\\
\hline
NowcastNet & 0.137 & 0.160 & 0.248 & 0.299 & 0.170 & \underline{0.207} & 0.128 & 0.139 & 0.097 & 0.106 & 0.040 & 0.047 \\
\hline
\textbf{TUPANN} & \textbf{0.169} & \textbf{0.187} & \textbf{0.267} & \textbf{0.312} & \textbf{0.189} & \textbf{0.211} & \textbf{0.177} & \textbf{0.177} & \textbf{0.135} & \textbf{0.141} & \textbf{0.079} & \textbf{0.094} \\
\textbf{GAN‑TUPANN} & \underline{0.152} & \underline{0.174} & \underline{0.252} & \underline{0.300} & \underline{0.180} & \textbf{0.211} & \underline{0.160} & \underline{0.164} & \underline{0.116} & \underline{0.128} & \underline{0.052} & \underline{0.066} \\
\hline

& \multicolumn{12}{|c}{\textbf{Manaus}\rule{0pt}{2.5ex}}\\
\hline
NowcastNet & 0.253 & 0.278 & 0.323 & \underline{0.366} & 0.296 & \textbf{0.324} & \underline{0.283} & \underline{0.303} & \underline{0.233} & 0.258 & 0.130 & 0.137 \\
\hline
\textbf{TUPANN} & \textbf{0.290} & \textbf{0.293} & \textbf{0.339} & \textbf{0.367} & \underline{0.316} & 0.321 & \textbf{0.315} & \underline{0.312} & \textbf{0.278} & \textbf{0.274} & \textbf{0.200} & \textbf{0.193} \\
\textbf{GAN‑TUPANN} & \underline{0.274} & \underline{0.285} & \underline{0.331} & 0.359 & \textbf{0.318} & \textbf{0.329} & \underline{0.310} & \textbf{0.313} & \underline{0.258} & \underline{0.267} & \underline{0.156} & \underline{0.156} \\
\hline

& \multicolumn{12}{|c}{\textbf{La Paz}\rule{0pt}{2.5ex}}\\
\hline
NowcastNet & 0.291 & 0.301 & 0.330 & \textbf{0.376} & 0.303 & 0.321 & 0.321 & 0.315 & 0.300 & 0.296 & 0.202 & 0.197 \\
\hline
\textbf{TUPANN} & \textbf{0.314} & \textbf{0.317} & \underline{0.336} & 0.363 & \underline{0.327} & \underline{0.323} & \textbf{0.350} & \textbf{0.340} & \textbf{0.327} & \textbf{0.322} & \textbf{0.232} & \textbf{0.239} \\
\textbf{GAN‑TUPANN} & \underline{0.306} & \underline{0.312} & \textbf{0.336} & \underline{0.367} & \textbf{0.327} & \textbf{0.333} & \underline{0.344} & \underline{0.336} & \underline{0.308} & \underline{0.305} & \underline{0.214} & \underline{0.218} \\
\bottomrule
\end{tabular}}
\label{tab:csi_gan_TUPANN}
\end{table*}

\subsubsection{Cross‑city generalization}
To assess generalization we train TUPANN on one city and evaluate on others.  \Cref{tab:cross-city} compares TUPANN trained on Rio de Janeiro (TUPANN–Rio) against models trained separately on each city.  In Manaus and La Paz, TUPANN–Rio yields lower CSI at all thresholds, as expected due to climate differences.  Surprisingly, in Miami the Rio‑trained model performs comparably or better at high thresholds, possibly because heavy‑rain events are rare in Miami (see \Cref{fig:prop_threshold_goes}) and the Rio model may bias towards such events. Overall, cross‑city degradation is modest (at most \(20\%\)) and TUPANN–Rio still outperforms or matches baselines trained on the target city, highlighting the transferability of the architecture.

\begin{table*}[ht]
\centering
\small
\renewcommand{\arraystretch}{1.2}
\caption{Cross‑city CSI comparison between TUPANN trained on each city and TUPANN trained on Rio de Janeiro (TUPANN–Rio).  Bold values denote the best, underlined values the second best.  Cross‑city performance declines in Manaus and La Paz but remains competitive; in Miami TUPANN–Rio improves high‑threshold scores.}
\resizebox{\textwidth}{!}{
\begin{tabular}{l|cc|cc|cc|cc|cc|cc}
\toprule
\multirow{2}{*}{Model}
 & \multicolumn{2}{c|}{$\textrm{CSI-M}\uparrow$}
 & \multicolumn{2}{c|}{$\textrm{CSI}_{4}\uparrow$}
 & \multicolumn{2}{c|}{$\textrm{CSI}_{8}\uparrow$}
 & \multicolumn{2}{c|}{$\textrm{CSI}_{16}\uparrow$}
 & \multicolumn{2}{c|}{$\textrm{CSI}_{32}\uparrow$}
 & \multicolumn{2}{c}{$\textrm{CSI}_{64}\uparrow$} \\
\cline{2-13}
 & POOL1 & POOL4 & POOL1 & POOL4 & POOL1 & POOL4 & POOL1 & POOL4 & POOL1 & POOL4 & POOL1 & POOL4 \\
\hline

& \multicolumn{12}{|c}{\textbf{Miami}\rule{0pt}{2.5ex}}\\
\hline
\textbf{TUPANN} & \textbf{0.169} & \textbf{0.187} & \textbf{0.267} & \textbf{0.312} & \textbf{0.189} & \textbf{0.211} & \underline{0.177} & \underline{0.177} & \underline{0.135} & \underline{0.141} & \underline{0.079} & \underline{0.094} \\
\textbf{TUPANN–Rio} & \underline{0.166} & \underline{0.185} & \underline{0.248} & \underline{0.289} & \underline{0.182} & \underline{0.199} & \textbf{0.178} & \textbf{0.182} & \textbf{0.138} & \textbf{0.152} & \textbf{0.085} & \textbf{0.103} \\
\hline

& \multicolumn{12}{|c}{\textbf{Manaus}\rule{0pt}{2.5ex}}\\
\hline
\textbf{TUPANN} & \textbf{0.290} & \textbf{0.293} & \textbf{0.339} & \textbf{0.367} & \textbf{0.316} & \textbf{0.321} & \textbf{0.315} & \textbf{0.312} & \textbf{0.278} & \textbf{0.274} & \textbf{0.200} & \textbf{0.193} \\
\textbf{TUPANN–Rio} & \underline{0.236} & \underline{0.249} & \underline{0.283} & \underline{0.319} & \underline{0.257} & \underline{0.276} & \underline{0.254} & \underline{0.264} & \underline{0.214} & \underline{0.221} & \underline{0.172} & \underline{0.167} \\
\hline

& \multicolumn{12}{|c}{\textbf{La Paz}\rule{0pt}{2.5ex}}\\
\hline
\textbf{TUPANN} & \textbf{0.314} & \textbf{0.317} & \textbf{0.336} & \textbf{0.363} & \textbf{0.327} & \textbf{0.323} & \textbf{0.350} & \textbf{0.340} & \textbf{0.327} & \textbf{0.322} & \textbf{0.232} & \textbf{0.239} \\
\textbf{TUPANN–Rio} & \underline{0.279} & \underline{0.288} & \underline{0.305} & \underline{0.339} & \underline{0.288} & \underline{0.295} & \underline{0.306} & \underline{0.304} & \underline{0.279} & \underline{0.279} & \underline{0.216} & \underline{0.221} \\
\bottomrule
\end{tabular}}
\label{tab:cross-city}
\end{table*}

\subsubsection{Multi‑city training}

Rather than training a model for each city, training on multiple cities can further improve skill.  \Cref{tab:multi-city} compares TUPANN trained separately on each city with a multi‑city model trained jointly on all regions (including Toronto, see \Cref{appendixA}).  The multi‑city model (TUPANN–Multicity) yields higher CSI across most thresholds and regions. Access to diverse climates helps the network learn more generalizable features, especially for extreme rainfall events.

\begin{table*}[ht]
\centering
\small
\renewcommand{\arraystretch}{1.2}
\caption{Comparison between single‑city TUPANN and a multi‑city version (TUPANN–Multicity) trained on all regions simultaneously.  Bold denotes the best, underlined the second best.  The multi‑city model improves performance in most cases, demonstrating the benefit of pooled training.}
\resizebox{\textwidth}{!}{
\begin{tabular}{l|cc|cc|cc|cc|cc|cc}
\toprule
\multirow{2}{*}{Model}
 & \multicolumn{2}{c|}{$\textrm{CSI-M}\uparrow$}
 & \multicolumn{2}{c|}{$\textrm{CSI}_{4}\uparrow$}
 & \multicolumn{2}{c|}{$\textrm{CSI}_{8}\uparrow$}
 & \multicolumn{2}{c|}{$\textrm{CSI}_{16}\uparrow$}
 & \multicolumn{2}{c|}{$\textrm{CSI}_{32}\uparrow$}
 & \multicolumn{2}{c}{$\textrm{CSI}_{64}\uparrow$} \\
\cline{2-13}
 & POOL1 & POOL4 & POOL1 & POOL4 & POOL1 & POOL4 & POOL1 & POOL4 & POOL1 & POOL4 & POOL1 & POOL4 \\
\hline

& \multicolumn{12}{|c}{\textbf{Rio de Janeiro}\rule{0pt}{2.5ex}}\\
\hline
\textbf{TUPANN} & \underline{0.259} & \underline{0.277} & \underline{0.330} & \underline{0.384} & \underline{0.289} & \underline{0.289} & \underline{0.330} & \underline{0.336} & \underline{0.274} & \underline{0.287} & \textbf{0.072} & \textbf{0.090} \\
\textbf{TUPANN–Multicity} & \textbf{0.271} & \textbf{0.286} & \textbf{0.337} & \textbf{0.390} & \textbf{0.303} & \textbf{0.297} & \textbf{0.353} & \textbf{0.353} & \textbf{0.293} & \textbf{0.304} & \underline{0.071} & \underline{0.086} \\
\hline

& \multicolumn{12}{|c}{\textbf{Miami}\rule{0pt}{2.5ex}}\\
\hline
\textbf{TUPANN} & \underline{0.169} & \underline{0.187} & \underline{0.267} & \underline{0.312} & \underline{0.189} & \textbf{0.211} & \underline{0.177} & \underline{0.177} & \underline{0.135} & \underline{0.141} & \underline{0.079} & \underline{0.094} \\
\textbf{TUPANN–Multicity} & \textbf{0.178} & \textbf{0.190} & \textbf{0.273} & \textbf{0.313} & \textbf{0.196} & \textbf{0.211} & \textbf{0.188} & \textbf{0.182} & \textbf{0.151} & \textbf{0.151} & \textbf{0.082} & \textbf{0.095} \\
\hline

& \multicolumn{12}{|c}{\textbf{Manaus}\rule{0pt}{2.5ex}}\\
\hline
\textbf{TUPANN} & \underline{0.290} & \underline{0.293} & \underline{0.339} & \underline{0.367} & \textbf{0.316} & \underline{0.321} & \textbf{0.315} & \underline{0.312} & \textbf{0.278} & \underline{0.274} & \underline{0.200} & \underline{0.193} \\
\textbf{TUPANN–Multicity} & \textbf{0.291} & \textbf{0.296} & \textbf{0.340} & \textbf{0.369} & \textbf{0.316} & \textbf{0.324} & \underline{0.314} & \textbf{0.314} & \textbf{0.278} & \textbf{0.275} & \textbf{0.208} & \textbf{0.198} \\
\hline

& \multicolumn{12}{|c}{\textbf{La Paz}\rule{0pt}{2.5ex}}\\
\hline
\textbf{TUPANN} & \underline{0.314} & \underline{0.317} & \underline{0.336} & \textbf{0.363} & \underline{0.327} & \underline{0.323} & \underline{0.350} & \underline{0.340} & \underline{0.327} & \underline{0.322} & \underline{0.232} & \underline{0.239} \\
\textbf{TUPANN–Multicity} & \textbf{0.324} & \textbf{0.325} & \textbf{0.339} & \textbf{0.363} & \textbf{0.334} & \textbf{0.327} & \textbf{0.359} & \textbf{0.348} & \textbf{0.335} & \textbf{0.331} & \textbf{0.251} & \textbf{0.254} \\
\bottomrule
\end{tabular}}
\label{tab:multi-city}
\end{table*}

\subsection{IMERG results}
\begin{figure}[ht]
    \centering
    \includegraphics{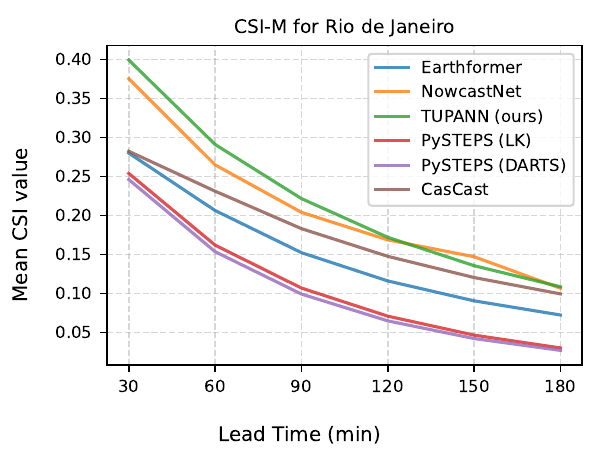}
    \caption{Mean CSI versus lead time for IMERG data in Rio de Janeiro.  TUPANN outperforms baselines at most lead times; NowcastNet overtakes slightly at 150 min but lags at shorter lead times}
    \label{fig:avg_csi_rio_imerg}
\end{figure}

\Cref{tab:metrics_rio_imerg} compares TUPANN to baselines on the IMERG dataset for Rio de Janeiro.  Without pooling (POOL1), TUPANN achieves the best CSI across all thresholds.  With pooling (POOL4), generative models (NowcastNet, GAN‑TUPANN) slightly outperform TUPANN at low thresholds but TUPANN remains competitive and leads at higher thresholds.  \Cref{fig:avg_csi_rio_imerg} plots CSI–M versus lead time, showing TUPANN’s superior performance at most lead times and small gaps only at 150 min.  These results demonstrate that TUPANN generalizes to coarser spatial resolution and longer latency datasets, where its physics-aligned architecture can become slightly less beneficial.

\begin{table*}[ht]
\centering
\caption{Aggregated CSI metrics for Rio de Janeiro using IMERG data.  Bold denotes the best, underlined the second best.  Without pooling TUPANN is clearly superior; with pooling, generative baselines perform slightly better at low thresholds, but TUPANN remains competitive overall.}
\renewcommand{\arraystretch}{1.3}
\resizebox{1.0\columnwidth}{!}{
\begin{tabular}{l|cc|cc|cc|cc|cc|cc}
\toprule
\multirow{2}{*}{Model}
 & \multicolumn{2}{c|}{$\textrm{CSI-M}\uparrow$}
 & \multicolumn{2}{c|}{$\textrm{CSI}_{4}\uparrow$}
 & \multicolumn{2}{c|}{$\textrm{CSI}_{8}\uparrow$}
 & \multicolumn{2}{c|}{$\textrm{CSI}_{16}\uparrow$}
 & \multicolumn{2}{c|}{$\textrm{CSI}_{32}\uparrow$}
 & \multicolumn{2}{c}{$\textrm{CSI}_{64}\uparrow$} \\
\cline{2-13}
 & POOL1 & POOL4 & POOL1 & POOL4 & POOL1 & POOL4 & POOL1 & POOL4 & POOL1 & POOL4 & POOL1 & POOL4 \\
\midrule
Earthformer & 0.153 & 0.141 & 0.361 & 0.343 & 0.270 & 0.243 & 0.130 & 0.114 & 0.006 & 0.006 & 0.000 & 0.000 \\
NowcastNet & \textbf{0.209} & \underline{0.271} & 0.387 & 0.453 & 0.322 & \underline{0.384} & 0.229 & \underline{0.304} & \textbf{0.103} & \textbf{0.201} & 0.001 & \textbf{0.013} \\
PySTEPS (LK) & 0.114 & 0.099 & 0.300 & 0.269 & 0.191 & 0.164 & 0.075 & 0.061 & 0.002 & 0.003 & 0.000 & 0.000 \\
PySTEPS (DARTS) & 0.107 & 0.096 & 0.287 & 0.262 & 0.179 & 0.157 & 0.069 & 0.057 & 0.002 & 0.003 & 0.000 & 0.000 \\
CasCast & 0.180 & 0.229 & 0.362 & 0.420 & 0.288 & 0.339 & 0.188 & 0.248 & 0.061 & 0.139 & 0.000 & 0.000 \\
\hline
\textbf{TUPANN (ours)} & \textbf{0.218} & 0.248 & \textbf{0.414} & \underline{0.454} & \textbf{0.344} & 0.379 & \textbf{0.241} & 0.280 & 0.087 & 0.123 & \textbf{0.005} & \underline{0.006} \\
\textbf{GAN‑TUPANN} & \underline{0.210} & \textbf{0.274} & \underline{0.391} & \textbf{0.461} & \underline{0.327} & \textbf{0.393} & \underline{0.234} & \textbf{0.313} & \underline{0.095} & \underline{0.199} & \underline{0.002} & 0.004 \\
\bottomrule
\end{tabular}}
\label{tab:metrics_rio_imerg}
\end{table*}

\subsection{Operational deployment}

Downloading and processing GOES-16 data and making predictions can be done in less than three minutes. Due to its low latency, this enables an efficient near real-time 10 to 180 minutes nowcast with 10 minutes temporal resolution. The whole pipeline, including the predictions generated by TUPANN, can be run in a machine with relatively modest requirements: 16GB of memory, GeForce RTX 3080 GPU. Predictions can also include the derived motion fields to help meteorologists interpret the results. This system is currently operational and has been deployed to aid Rio de Janeiro's Instituto Estadual do Ambiente (INEA) agency to prepare for floods and other extreme precipitation disasters. Due to TUPANN's reliance on widely available geostationary satellite data, we expect that this solution can be broadened to other regions as well.

\section{Limitations and future work}\label{sec:limitations}

Our study has important limitations.  First, although TUPANN may use globally available satellite products, it currently relies on GOES‑16 coverage for real‑time nowcasting.  The architecture should be retrained and validated on other geostationary satellites (e.g., Himawari, Meteosat) to ensure generalization across platforms. Second, the optical‑flow supervision requires additional computation and may not perfectly capture complex convection dynamics; using the proposed scheme with additional covariates could yield further improvements under a more detailed physical modeling. Third, CSI and HSS values decline at longer lead times (>2h), reflecting inherent predictability limits; integrating uncertainty quantification and ensemble approaches may better characterize forecast skill. Finally, GAN‑based enhancements improve visual realism but degrade or inconsistently affect skill metrics; stabilizing adversarial training and assessing perceptual quality remain open challenges.

Future work includes coupling TUPANN with probabilistic post‑processing to provide calibrated uncertainty estimates, extending the model to include additional inputs such as cloud imagery or microwave precipitation retrievals, and investigating transfer learning across continents.

\section{Conclusions}\label{sec:conclusion}

We have presented TUPANN, a physically aligned neural network for precipitation nowcasting using satellite imagery.  TUPANN’s modular design --- combining a variational encoder–decoder supervised by optical flow, and a transformer capable of evolving the latent representation according to physical constraints --- yields interpretable motion fields and competitive forecast skill. Extensive experiments on GOES‑16 and IMERG data across four climates show that TUPANN matches or surpasses state‑of‑the‑art baselines, particularly at high precipitation thresholds. Training on multiple cities improves performance and cross‑city evaluations reveal modest degradation, highlighting the model’s transferability.  With its low latency and reliance on globally available satellites, TUPANN supports equitable access to short‑term rainfall forecasts and provides a foundation for operational applications in radar‑sparse regions.

\paragraph{Acknowledgements} The authors thank Adriana Monteiro, João Vitor Romano, Lucas Nissenbaum,  and Thiago Ramos as well as the help and support from Google, in particular, Shreya Agrawal, Boris Babenko and Samier Merchant.

\paragraph{Funding Statement} A.C. was supported by a “FAPERJ Nota 10" grant (SEI-260003/004731/2025) from Fundação de Amparo à Pesquisa do Estado do Rio de Janeiro (FAPERJ). M.P. and L.V. were supported by scholarships from Coordenação de Aperfeiçoamento de Pessoal de Nível Superior (CAPES). P.O. was supported by grant SEI-260003/001545/2022 from FAPERJ.

\paragraph{Competing Interests} The authors declare no competing interests.

\paragraph{Data Availability Statement} GOES‑16 RRQPE data were obtained from the NOAA GOES‑R Product Distribution and Access (PDA) system.  IMERG Final Run data were obtained from the NASA GPM data portal.  Processed datasets, event splits and TUPANN code are available at \url{https://github.com/acataos/tupann}.

\paragraph{Ethical Standards} The research meets all ethical guidelines, including adherence to the legal requirements of Brazil and the United States.

\paragraph{Author Contributions} Conceptualization: A.C., M.P., L.V., P.O.; Methodology: A.C., M.P., L.V.; Data curation and visualization: A.C.; Writing—original draft: A.C., M.P., L.V.; Writing—review and editing: A.C., M.P., L.V., P.O..  All authors approved the final manuscript.

\bibliographystyle{apalike}
\bibliography{references}

\newpage

\begin{appendix}

\section{Additional results}\label[appendix]{appendixA}

Some additional results related to metrics and evaluation are presented below, such as HSS and CSI scores for the city of Toronto. Prediction samples in both GOES-16 and IMERG data are also included.

\subsection{Model Hyperparameters}

The hyperparameters for Earthformer and TUPANN (both VED and MaxViT) are shown below. Hyperparameters  Evolution Network/NowcastNet were the ones selected in the original paper \citep{zhang2023}. 
Since CasCast uses a Denoising Transformer (DiT) model trained with a different image size (384x384), slight adaptations were made to support the 256x256 images used in this work. The hyperparameters modified were input\_size to 32 and hidden\_size to 512; other hyperparameters were chosen as in their original paper \citep{cascast}. All remaining hyperparameters and early stopping criteria were selected by maximizing the mean CSI value in the validation set during training.

\begin{table}[h]  %
\centering
\caption{Model hyperparameters for VED, Earthformer, and MaxViT. Architectural and loss parameters are detailed in \Cref{sec:methods}.}
\begin{subtable}[t]{0.3\textwidth}
\centering
\caption{VED}
\begin{tabular}{l c}
\toprule
Parameter & Value \\
\midrule
batch\_size & 8 \\
learning\_rate & 0.0001 \\
channels & 128 \\
embed\_dim & 4 \\
reduc\_factor & 4 \\
$\lambda_{\textrm{cos}}$ & 0.00165 \\
$\lambda_{\textrm{KL}}$ & 1.0e-06 \\
$\lambda_{\textrm{motion}}$ & 0.0033 \\
$\lambda_{\textrm{int}}$ & 0.995 \\
dropout & 0.2 \\
\bottomrule
\end{tabular}
\end{subtable}
\hfill
\begin{subtable}[t]{0.3\textwidth}
\centering
\caption{Earthformer}
\begin{tabular}{l c}
\toprule
Parameter & Value \\
\midrule
batch\_size & 4 \\
learning\_rate & 0.0001 \\
num\_global\_vectors & 6 \\
num\_heads & 2 \\
base\_units & 64 \\
\bottomrule
\end{tabular}
\end{subtable}
\hfill
\begin{subtable}[t]{0.3\textwidth}
\centering
\caption{MaxViT}
\begin{tabular}{l c}
\toprule
Parameter & Value \\
\midrule
batch\_size & 8 \\
learning\_rate & 0.0001 \\
MaxViT\_depth & 4 \\
MaxViT\_dim & 64 \\
\bottomrule
\end{tabular}
\end{subtable}
\end{table}

\subsection{HSS results}
\Cref{tab:hss_final} summarizes aggregated HSS metrics for the four cities using GOES‑16 data.  TUPANN achieves the best scores for high thresholds and remains competitive for lower thresholds, mirroring the CSI behaviour.

\begin{table*}[h]
\centering
\caption{Aggregated HSS metrics for GOES‑16 data across cities.  Bold values denote the best, underlined values the second best.  TUPANN excels at high thresholds and is second or first at lower thresholds.}
\renewcommand{\arraystretch}{1.3}
\resizebox{0.75\textwidth}{!}{
\begin{tabular}{l|c|c|c|c|c|c}
\hline
{Model}
 & {$\textrm{HSS-M}\uparrow$}
 & {$\textrm{HSS}_{4}\uparrow$}
 & {$\textrm{HSS}_{8}\uparrow$}
 & {$\textrm{HSS}_{16}\uparrow$}
 & {$\textrm{HSS}_{32}\uparrow$}
 & {$\textrm{HSS}_{64}\uparrow$} \\
\hline

& \multicolumn{6}{|c}{\textbf{Rio de Janeiro}\rule{0pt}{2.5ex}}\\
\hline
Earthformer & 0.360 & \underline{0.473} & \underline{0.438} & \underline{0.488} & 0.382 & 0.017 \\
NowcastNet & \underline{0.373} & 0.454 & 0.429 & 0.478 & \underline{0.394} & \underline{0.112} \\
PySTEPS (LK) & 0.266 & 0.369 & 0.341 & 0.365 & 0.247 & 0.010 \\
PySTEPS (DARTS) & 0.268 & 0.356 & 0.347 & 0.369 & 0.244 & 0.025 \\
CasCast & 0.266 & 0.438 & 0.321 & 0.271 & 0.270 & 0.032 \\
\hline
\textbf{TUPANN (ours)} & \textbf{0.393} & \textbf{0.473} & \textbf{0.439} & \textbf{0.492} & \textbf{0.428} & \textbf{0.135} \\
\hline

& \multicolumn{6}{|c}{\textbf{Miami}\rule{0pt}{2.5ex}}\\
\hline
Earthformer & 0.230 & \textbf{0.412} & 0.299 & \underline{0.265} & 0.176 & 0.000 \\
NowcastNet & 0.224 & 0.368 & 0.277 & 0.223 & 0.177 & 0.076 \\
PySTEPS (LK) & 0.195 & 0.299 & 0.227 & 0.213 & 0.147 & \underline{0.087} \\
PySTEPS (DARTS) & 0.192 & 0.301 & 0.231 & 0.210 & 0.135 & 0.083 \\
CasCast & \underline{0.237} & 0.388 & \underline{0.304} & 0.247 & \underline{0.208} & 0.039 \\
\hline
\textbf{TUPANN (ours)} & \textbf{0.277} & \underline{0.398} & \textbf{0.309} & \textbf{0.298} & \textbf{0.237} & \textbf{0.146} \\
\hline

& \multicolumn{6}{|c}{\textbf{Manaus}\rule{0pt}{2.5ex}}\\
\hline
Earthformer & \underline{0.417} & \textbf{0.502} & \textbf{0.476} & \textbf{0.472} & \underline{0.414} & 0.220 \\
NowcastNet & 0.384 & 0.457 & 0.437 & 0.429 & 0.372 & \underline{0.228} \\
PySTEPS (LK) & 0.319 & 0.385 & 0.369 & 0.349 & 0.271 & 0.222 \\
PySTEPS (DARTS) & 0.314 & 0.386 & 0.371 & 0.350 & 0.269 & 0.195 \\
CasCast & 0.401 & \underline{0.486} & 0.447 & 0.444 & 0.406 & 0.222 \\
\hline
\textbf{TUPANN (ours)} & \textbf{0.435} & 0.479 & \underline{0.464} & \underline{0.469} & \textbf{0.430} & \textbf{0.333} \\
\hline

& \multicolumn{6}{|c}{\textbf{La Paz}\rule{0pt}{2.5ex}}\\
\hline
Earthformer & \underline{0.454} & \textbf{0.487} & \textbf{0.486} & \textbf{0.523} & \underline{0.485} & 0.285 \\
NowcastNet & 0.439 & 0.472 & 0.451 & 0.479 & 0.458 & \underline{0.335} \\
PySTEPS (LK) & 0.340 & 0.378 & 0.381 & 0.390 & 0.326 & 0.223 \\
PySTEPS (DARTS) & 0.356 & 0.398 & 0.405 & 0.412 & 0.339 & 0.225 \\
CasCast & 0.351 & 0.441 & 0.381 & 0.381 & 0.370 & 0.183 \\
\hline
\textbf{TUPANN (ours)} & \textbf{0.468} & \underline{0.482} & \underline{0.481} & \underline{0.512} & \textbf{0.490} & \textbf{0.376} \\
\bottomrule
\end{tabular}}
\label{tab:hss_final}
\end{table*}

\subsection{Visual inspection of model predictions}

\begin{figure}[htpb] 
    \centering
    \includegraphics[scale=0.45]{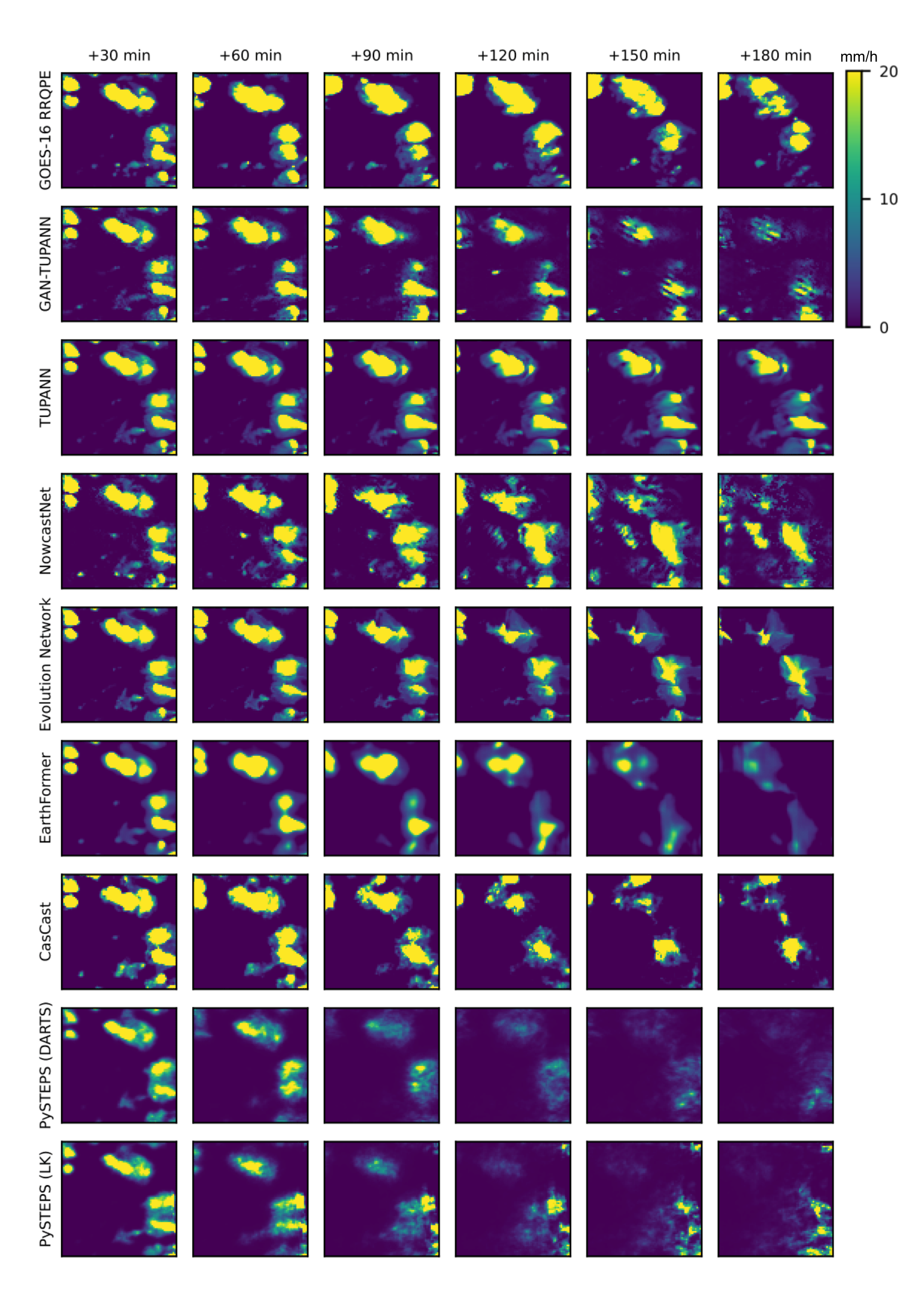}
    \caption{Predictions up to 3h ahead for a rain event in Manaus, starting at 2021‑11‑20 22:00 UTC. Rows show the ground truth, TUPANN, and other models; columns represent lead times. TUPANN displays greater skill in predicting the movement and intensity of the rain event, while generative models produce visually sharper but less accurate predictions}
    \label{fig:prediction_example_big}
\end{figure}

\begin{figure}[htpb] 
    \centering
    \includegraphics[scale=0.45]{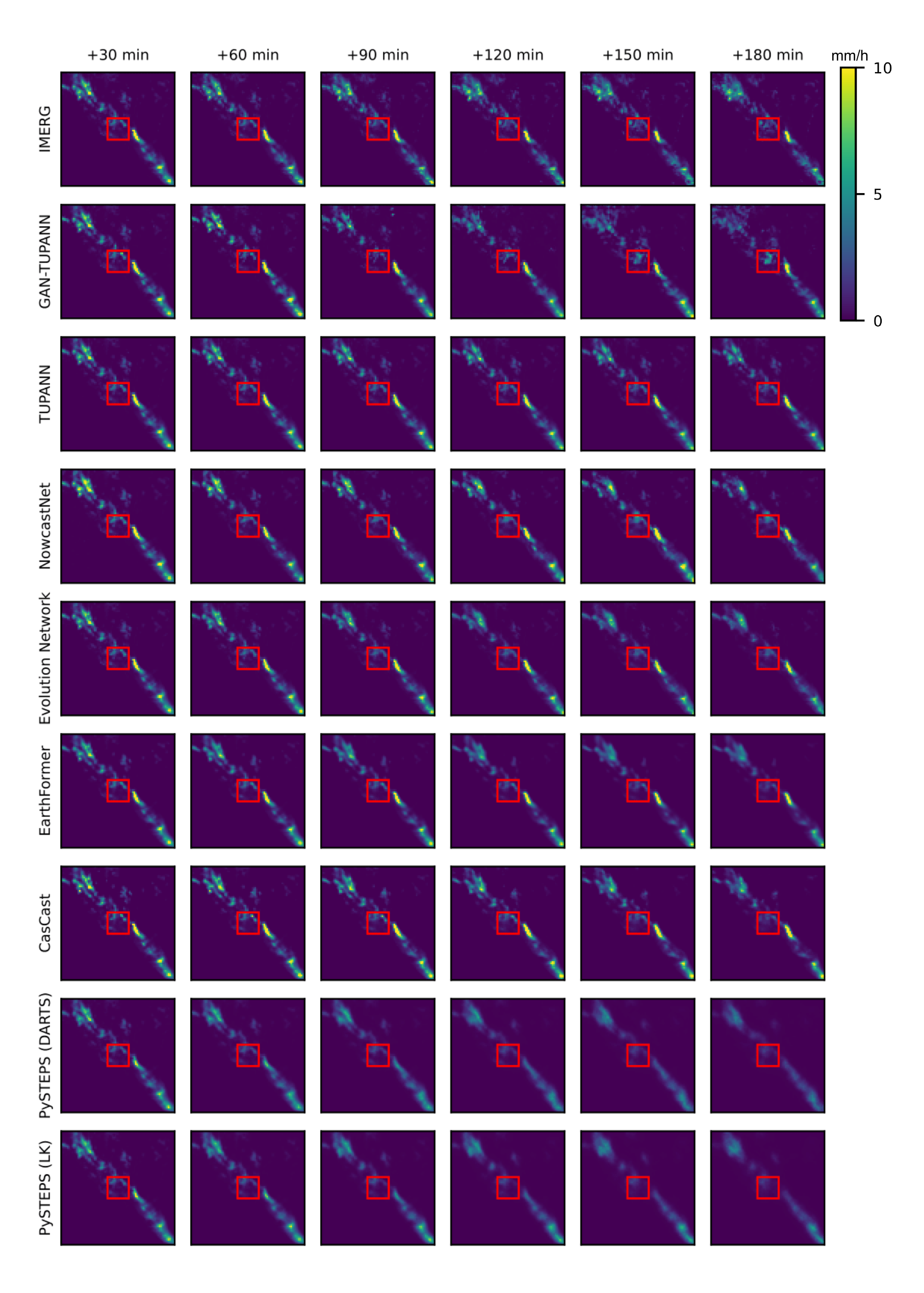}
    \caption{Prediction sample from the IMERG test dataset centered in Rio de Janeiro, starting from 2023-01-08 07:30 UTC. The red square represents the same area as in the GOES-16 figures, demonstrating the difference in spatial resolution}
    \label{fig:prediction_example_big_imerg}
\end{figure}

\Cref{fig:prediction_example_big} includes predictions for several models at a given moment in time. While TUPANN and EarthFormer show relatively blurred predictions, note NowcastNet and CasCast add several artifacts to its predictions. Overall, TUPANN achieves a reasonable trade-off between good evaluation metrics and reasonable precipitation plots. \Cref{fig:prediction_example_big_imerg} also includes precipitation plots for the IMERG dataset, where the temporal and spatial resolution of the image is coarser.

\subsection{Results for Toronto}
For completeness \Cref{tab:toronto} reports CSI scores for the city of Toronto using GOES‑16 data. This city was excluded from the main text once it did not present as many extreme precipitation events as the other selected alternatives. The multi‑city TUPANN model performs best across most thresholds, while the single‑city TUPANN model ranks second.  Although extreme precipitation events are rare in Toronto, these results demonstrate that TUPANN generalizes to additional regions.

\begin{table*}[h]
\centering
\caption{Aggregated CSI metrics for Toronto with GOES‑16 data.  Bold denotes the best, underlined the second best.}
\renewcommand{\arraystretch}{1.3}
\resizebox{\textwidth}{!}{
\begin{tabular}{l|cc|cc|cc|cc|cc|cc}
\toprule
\multirow{2}{*}{Model}
 & \multicolumn{2}{c|}{$\textrm{CSI-M}\uparrow$}
 & \multicolumn{2}{c|}{$\textrm{CSI}_{4}\uparrow$}
 & \multicolumn{2}{c|}{$\textrm{CSI}_{8}\uparrow$}
 & \multicolumn{2}{c|}{$\textrm{CSI}_{16}\uparrow$}
 & \multicolumn{2}{c|}{$\textrm{CSI}_{32}\uparrow$}
 & \multicolumn{2}{c}{$\textrm{CSI}_{64}\uparrow$} \\
\cline{2-13}
 & POOL1 & POOL4 & POOL1 & POOL4 & POOL1 & POOL4 & POOL1 & POOL4 & POOL1 & POOL4 & POOL1 & POOL4 \\
\midrule
Earthformer & 0.209 & 0.187 & 0.288 & 0.280 & 0.233 & 0.203 & 0.154 & 0.133 & 0.102 & 0.084 & 0.000 & 0.000 \\
NowcastNet & 0.206 & \underline{0.216} & 0.278 & 0.294 & 0.225 & 0.238 & 0.151 & 0.161 & 0.100 & 0.105 & 0.000 & 0.000 \\
PySTEPS (LK) & 0.179 & 0.173 & 0.250 & 0.240 & 0.200 & 0.186 & 0.130 & 0.120 & 0.078 & 0.070 & 0.000 & 0.000 \\
PySTEPS (DARTS) & 0.181 & 0.174 & 0.252 & 0.242 & 0.203 & 0.188 & 0.133 & 0.121 & 0.080 & 0.071 & 0.000 & 0.000 \\
CasCast & 0.192 & 0.203 & 0.270 & \underline{0.286} & 0.217 & \underline{0.231} & 0.145 & \underline{0.158} & 0.096 & \underline{0.105} & 0.000 & 0.000 \\
\hline
\textbf{TUPANN} & \underline{0.219} & \underline{0.222} & \underline{0.295} & 0.295 & \underline{0.239} & \underline{0.231} & \underline{0.159} & \underline{0.158} & \underline{0.106} & \underline{0.105} & 0.000 & 0.000 \\
\textbf{TUPANN–Multicity} & \textbf{0.229} & \textbf{0.230} & \textbf{0.305} & \textbf{0.298} & \textbf{0.250} & \textbf{0.241} & \textbf{0.170} & \textbf{0.165} & \textbf{0.114} & \textbf{0.107} & 0.000 & 0.000 \\
\bottomrule
\end{tabular}}
\label{tab:toronto}
\end{table*}

\subsection{Rain events dataset selection}\label{sec:rain_events}
The datasets used to train and evaluate our models comprise a subsample of rainy windows drawn from either the GOES-16 RRQPE or the IMERG products. We define a rainy window as follows. For each 10-min timestamp \(t\) from 2020-01-01 00:00~UTC to 2023-12-31 23:50~UTC, we (i) form a symmetric 60-min window \([t-30\,\text{min},\, t+30\,\text{min}]\); (ii) compute the spatiotemporal precipitation accumulation over that window (10-min steps over all grid points); and (iii) if the accumulation exceeds a threshold \(\tau\), label the larger window \( [t-4\,\text{hours},t+4\,\text{hours}] \) as a rainy window. Finally, we merge rainy windows that intersect. The threshold \(\tau=120{,}000\) was chosen empirically to balance excluding near-dry periods against obtaining a dataset large enough for effective learning. Using a symmetric \(\pm 4\)-hour window ensures that events include both onset and dissipation phases (from no precipitation to mild or heavy precipitation and back).

\end{appendix}
\end{document}